\documentclass{article}

\pdfoutput=1
\usepackage{fullpage}
\usepackage[utf8]{inputenc}
\usepackage{graphicx}
\usepackage{amssymb}
\usepackage{amsmath}
\usepackage{amsfonts}
\usepackage{color}
\usepackage[pdftex]{hyperref}
\usepackage[english]{babel}
\usepackage{subfigure}
\usepackage{algorithm}
\usepackage{algorithmic}
\usepackage{multicol}
\usepackage{multirow}
\usepackage{tabularx,colortbl}
\usepackage{textcomp}
\usepackage{booktabs}
\usepackage{longtable}
\usepackage{array,longtable,multirow}
\usepackage[sort&compress,numbers]{natbib}
\usepackage{authblk}

\definecolor{Orange}{rgb}{1,0.64,0}
\definecolor{lgray}{rgb}{0.9,0.9,0.9}

\newcommand{\argmin}{\operatornamewithlimits{arg\ min}}

\makeindex

\begin{document}

\title{Modeling and Recognition of Smart Grid Faults by a Combined Approach of Dissimilarity Learning and One-Class Classification}

\author[1]{Enrico {De Santis}\thanks{enrico.desantis@uniroma1.it}\thanks{Corresponding author}}
\author[2]{Lorenzo Livi\thanks{llivi@scs.ryerson.ca}\thanks{Corresponding author}}
\author[2]{Alireza Sadeghian\thanks{asadeghi@ryerson.ca}}
\author[1]{Antonello Rizzi\thanks{antonello.rizzi@uniroma1.it}}
\affil[1]{Dept. of Information Engineering, Electronics, and Telecommunications, SAPIENZA University of Rome, Via Eudossiana 18, 00184 Rome, Italy}
\affil[2]{Dept. of Computer Science, Ryerson University, 350 Victoria Street, Toronto, ON M5B 2K3, Canada}
\renewcommand\Authands{, and }
\providecommand{\keywords}[1]{\textbf{\textit{Index terms---}} #1}

\maketitle

\begin{abstract}
Detecting faults in electrical power grids is of paramount importance, either from the electricity operator and consumer viewpoints.
Modern electric power grids (smart grids) are equipped with smart sensors that allow to gather real-time information regarding the physical status of all the component elements belonging to the whole infrastructure (e.g., cables and related insulation, transformers, breakers and so on).
In real-world smart grid systems, usually, additional information that are related to the operational status of the grid itself are collected such as meteorological information.
Designing a suitable recognition (discrimination) model of faults in a real-world smart grid system is hence a challenging task.
This follows from the heterogeneity of the information that actually determine a typical fault condition.
The second point is that, for synthesizing a recognition model, in practice only the conditions of observed faults are usually meaningful. Therefore, a suitable recognition model should be synthesized by making use of the observed fault conditions only.
In this paper, we deal with the problem of modeling and recognizing faults in a real-world smart grid system, which supplies the entire city of Rome, Italy.
Recognition of faults is addressed by following a combined approach of multiple dissimilarity measures customization and one-class classification techniques.
We provide here an in-depth study related to the available data and to the models synthesized by the proposed one-class classifier.
We offer also a comprehensive analysis of the fault recognition results by exploiting a fuzzy set based reliability decision rule.\\
\end{abstract}

\section{Introduction}
\label{sec:intro}

There are many possible definitions for a Smart Grid (SG). The SG European Technology Platform defines \cite{smartgrids_ETP} a SG as an ``electricity network that can intelligently integrate the actions of all the connected users, generators, consumers and those that do both, in order to efficiently deliver sustainable economic and secure electricity supply.''
A SG employs innovative products and services together with intelligent monitoring, control, communication, and self-healing technologies in order to (i) facilitate the connection and operation of generators of all sizes and technologies; (ii) allow consumers to play an active role in optimizing the operation of the system; (iii) significantly reduce the environmental impact of the whole electricity supply system; (iv) preserve or improve the level of system reliability, quality of service, and security.
SGs can be considered as an ``evolution'' rather than a ``revolution'' of the existing energy networks \cite{Energy_Information_Administration}. The evolution is leaded by the symbiotic exchange between power grid technologies and the Information and Communication Technologies (ICT). ICT provide instruments, such as \textit{Smart Sensors} (SS), to monitor the network status, wired and wireless communication network to collect and transport data, and powerful computational architectures for data processing.
A SG can be framed as a complex non-linear and time-varying system \cite{dorfler2013synchronization,amin2005toward,5535240,Venayagamoorthy__2009,mei_power_2011,machowski2011power}, where heterogeneous elements, including exogenous factors, are extremely interconnected through the exchange of both energy and information.
Computational Intelligence (CI) techniques offer sound modeling and algorithmic solutions in the SG context \cite{5179088,5952102,abido2014computational}.
Well-known CI techniques adopted in the SG context include approximate dynamic programming \cite{5768096}, neural networks and fuzzy inference systems for prediction and control \cite{6111641,molderink2010management}, and swarm intelligence and evolutionary computation for optimization problems \cite{de2013genetic,6608435,abdelaziz2009distribution}.

An important key issue of SGs is the design of a Decision Support System (DSS), which is an expert system that provides decision support for the commanding and dispatching systems of the power grid.
Such a system analyzes the risk for damage of crucial equipments, assesses the power grid security, forecasts and provides warnings about the magnitude and location of possible faults, and timely broadcasts the early-warning signals through suitable communication networks \cite{mei_power_2011}.
The information provided by the DSS can be used for Condition Based Maintenance (CBM) in the power grid \cite{Raheja2006-J-IJPR}.
CBM is defined as ``a philosophy that posits repair or replacement decisions on the current or future condition of assets''.
The objective of CBM is thus to minimize the total cost of inspection and repair by collecting and interpreting (heterogeneous) data related to the operating condition of critical components.
Through the use of CBM, advanced SS technology has the potential to help utilities to improve the power grid reliability by avoiding unexpected outages.
A discussion on how the changes in modern power grids have affected the maintenance procedures can be found in \cite{1600559}; the importance of modern diagnostic techniques is treated in \cite{6039785}.

Collecting heterogeneous measurements in modern SG systems is of paramount importance. As an instance, the available measurements can be used for dealing with various important pattern recognition and data mining problems on SGs, such as fault recognition \cite{6175733,Zhang2011791,Saha2011887}.
On the basis of the data at hand, different problem types could be formulated.
In \cite{1645199} the authors have established a relationship between environmental features and fault causes. A fault cause classifier based on the linear discriminant analysis (LDA) is proposed in \cite{5275689}. Information regarding weather conditions, longitude-latitude information, and measurements of physical quantities (e.g., currents and voltages) related to the power grid have been taken into account. In \cite{5156572}, the authors proposed a system based on LDA, which processes phasor measurement unit data, with the aim of recognizing and locating faults on power lines.
As concerns fault diagnosis in power grids, in \cite{5234528} is proposed a Support Vector Machine (SVM) based method to perform the recognition of faults related to high-voltage transmission lines.
The One-Class Quarter-Sphere SVM algorithm is proposed \cite{6477812} for faults classification in the power grid. The reported experimental evaluation is however performed on synthetically generated data only.

ACEA is the electricity distribution company managing the electrical network feeding the entire city of Rome. In this paper, we extend our previous work \cite{enrico_occ} on the problem of modeling and recognizing faults in the real-world SG system of ACEA by introducing several improvements.
Initially, we introduce the application's context and the approach followed to implement the one-class classification system used to recognize conditions of fault.
Since the available ACEA data is highly structured (i.e., it is formed by several heterogeneous information), we designed a dedicated one-class classifier (OCC) that is suitable for the specific application at hand.
The first herein presented improvement consists in equipping the designed OCC with the capability of producing also \textit{soft} output decisions. This is implemented by interpreting the decision regions synthesized by the classifier as fuzzy sets with suitable membership functions \cite{Livi_ga_2013,eocc__arxiv,pedrycz1998introduction}.
This fact allows us to provide also a measure of reliability concerning the already implemented hard classification mechanism.
As concerns the experiments, we provide (i) several evaluations of the recognition systems on either ad-hoc synthetic and ACEA datasets, (ii) a comparison on some well-known UCI datasets \cite{Bache+Lichman:2013} with other state of the art OCCs, and finally (iii) a more in depth analysis of the informativeness of the solutions found by the proposed OCC on the ACEA data.

The paper is structured as follows.
We offer a brief review on the one-class classification setting in Sec. \ref{sec:occ_prob}.
In Sec. \ref{sec:SG_proj} we provide a short overview about the main project where this study is collocated.
In Sec. \ref{sec:SG}, we describe the technical details of the considered SG.
Sec. \ref{sec:system} introduces the fault recognition system that we designed for the specific application at hand.
In this section, we describe (i) the representation of a fault pattern instance and (ii) the computational system as a whole highlighting also the new contributions introduced here in this paper.
In Sec. \ref{sec:experiments} we discuss the experiments.
Finally, in Sec. \ref{sec:conclusions} we draw our conclusions.

\section{Brief Overview on the One-class Classification Problem}
\label{sec:occ_prob}

The one-class classification problem can be considered as a particular instance of a standard $n$-class classification problem, where, during the training stage, patterns belonging only to a specific class are available. Such patterns are usually termed \textit{target} or \textit{positive} patterns.
This particular scenario covers several interesting real-world situations \cite{Ding2014313,one-class_survey__2010,oilspill__2010,Kemmler201329,enrico_occ,Utkin:2012:FOC:2213741.2433967,NIPS2002_2163}.
Practically, OCCs define a decision rule on the basis of a model that is able to describe suitable \textit{boundaries} pertaining the target patterns. Such boundaries define the decision regions/surface of the classifier. The aim is to synthesize effective models such that target patterns are recognized while non-target patterns are rejected.

\citet{one-class_survey__2010} provided a recent survey on the subject of one-class classification.
One important class of OCCs has been elaborated from the well-known SVM \cite{Tax19991191,SchWilSmoShaetal00,Wang2013875}.
\citet{Tax19991191} defined well-known system called Support Vector Data Description (SVDD).
The classification model is defined in terms of hyper-spheres, which are placed over the training set through an SVM-like optimization problem (the minimization of the sphere radiuses is enforced).
SVDD can be extended to different input domains by defining suitable positive definite kernel functions.

\citet{SchWilSmoShaetal00} proposed an alternative approach to SVDD that employs a hyperplane, like in the conventional SVM case.
The hyperplane is synthesized towards the aim of separating the region of the input space containing (target) patterns form the region containing no data. Also this approach has the capability of using kernel functions.
Other more recent approaches include algorithms based on the minimum spanning tree \cite{Juszczak20091859} and on Gaussian processes \cite{Kemmler2013}.

\section{An Overview on the ACEA Smart Grid Project}
\label{sec:SG_proj}

\begin{figure}[ht!]
 \centering
 \includegraphics[viewport=0 0 855 347,scale=0.4,keepaspectratio=true]{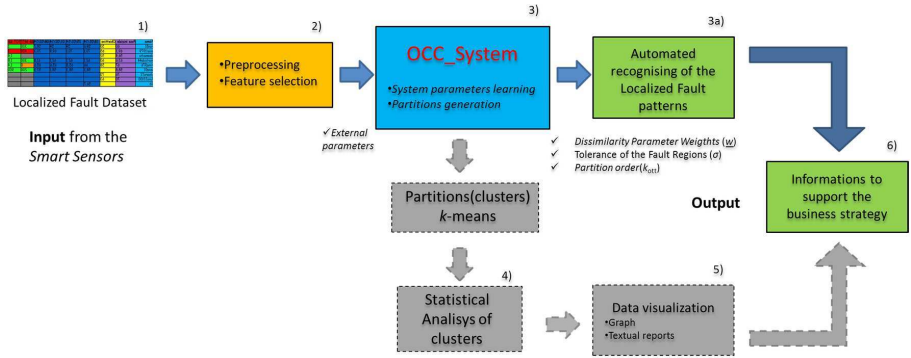}
 \caption{Process flow diagram describing the``ACEA Smart Grid Project''.}
 \label{fig:Process_flow}
\end{figure}

The following work represents a branch of a larger project, namely ``ACEA Smart Grid Pilot Project'' \cite{ACEA_SG_Pilot_Proj}.
The project aims to develop an automated recognition tool of fault states in the ACEA power grid.
In addition, the tool is designed to offer also diagnostic features, allowing the characterization of the power grid status during fault events.
The process flow diagram depicted in Fig. \ref{fig:Process_flow} shows the overall system and how raw data coming from the SS are transformed into meaningful information in order to support business strategies.
To obtain the dataset herein elaborated, a preliminary preprocessing stage, operated together with the ACEA experts, is performed. The dataset is then used as input for the herein presented OCC, which by means of an evolutionary strategy is in charge of learning typical situations of faults.
Clustering techniques are used to define the model of the proposed OCC. The synthesized partition is used also for post-processing purposes, such as data analysis and visualization. Those last two post-processing stages, belonging to the work packages set of the overall project, are not discussed in this paper.

\section{The Considered Power Grid}
\label{sec:SG}

The considered grid is constituted of backbones of uniform section, exerting radially with the possibility of counter-supply if a branch is out of order. Each backbone of the power grid is supplied by two distinct Primary Stations (PS) and each half-line is protected against faults through the breakers.
The Medium Voltage (MV) power grid consists in lines (feeders) in which the nominal voltage is 20 kV, with the presence of few ``legacy'' lines that still work at 8.4 kV. The MV part of the network covers 10,490 km while the Low Voltage (LV) section covers 11.120 km. 
Cables can be on air or underground and their sections can vary along the backbone with the presence of bottlenecks.
The MV section has 1.565 lines in service and it is supplied with 76 PSs, while LV section is supplied with 13.292 Secondary Stations (SS). Each MV line feeds a given number of secondary stations (MV-LV transformers), each one provided of two breakers, so that the substation can be isolated from the feeder in case of fault. These breakers on the medium voltage busbar are used to assure the radiality condition of the network, stating that each substation must be fed by only one PS.

We deal with the problem of modeling and recognizing a particular type of faults, which are commonly termed as localized faults (LFs) \cite{Acea_Road_mapr_2010}. Before providing a precise definition for a LF, it is important to discriminate among \textit{outages} and \textit{faults}, according to the CEI 5160 normative \cite{Doe:2013:Online}.
An outage (i.e., an interruption of the service) is the condition in which the voltage on the access point to the electrical energy of a user is less than 5\% of the declared voltage on all phases of supply \cite{Acea_Road_mapr_2010}.
Three types of outage are considered: (i) long, if the duration is more than three minutes (long outages); (ii) short, if the duration is more than one second and less than three minutes (short outages); (iii) transient, if the duration does not exceed one second (transient outages).

A fault, instead, is related to the failure of the electrical insulation (e.g., insulation of cables) that compromises the correct functioning of the grid.
Therefore, a LF is effectively a fault in which a physical element of the grid is permanently damaged causing long outages.

\section{The Proposed One-class Classification System for Smart Grid Fault Detection}
\label{sec:system}

\subsection{Representation of a Fault Pattern}
\label{sec:pattern_representation}

Instances of fault patterns (FPs) describing LFs occurred in the SG have been elaborated from a historical database provided by ACEA. The considered period spans across 2009--2012. During this time period the electrical network was not provided of any ICT infrastructure to record in a proper data base faults events, neither normal working conditions. Faults were recorded by manual entry following a given protocol. As a consequence these fault records are the only available data for this study.
Faults are characterized by heterogeneous data, including weather conditions, spatio-temporal data (i.e., longitude-latitude pairs and timestamps), physical data related to the state of power grid and its electric equipments (e.g., measured currents and voltages), and finally meteorological data.
As a consequence, a FP is actually defined by features of different nature, containing categorical (nominal) data, quantitative data (i.e., data having a well-defined metric), and also Time Series (TS) describing short outages occurred before a LF.
A detailed description of the considered features characterizing a FP is provided in Tab. \ref{tab:features_description}.
\begin{table*}[htbph!]\scriptsize
\centering
\caption{Description of the considered features describing a FP instance.}
\begin{tabular} {|p{4.5cm}|p{3.5cm}|p{5cm}|}  
\hline 
\rowcolor{lgray} \bf{Feature} &   \bf{Data typology and features space label} &   \bf{Description}\\
\hline
\bf{Day start} & Quantitative (Integer) $\mathcal{F}_{D}$  & Day in which the LF was detected \\
 \hline
\bf{Time start} & Quantitative (Integer) $\mathcal{F}_{T}$ &  Time stamp (minutes) in which the LF was detected\\
 \hline
\bf{Location element}& Categorical (String) $\mathcal{F}_{1}^{C}$ & Element positioning (aerial or underground) \\
 \hline
\bf{Material} & Categorical (String) $\mathcal{F}_{2}^{C}$ & Constituent material element (CU, AL) \\
\hline
\bf{Primary station fault distance} & Quantitative (Real) $\mathcal{F}_{1}^{Q}$  & Distance between the primary station and the geographical location of the LF \\
\hline
\bf{Median point} & Quantitative (Real) $\mathcal{F}_{2}^{Q}$ & Fault location calculated as median point between two secondary stations \\
\hline
\bf{\# Secondary Stations (SS)} & Quantitative (Real) $\mathcal{F}_{3}^{Q}$  & Number of out of service secondary stations after the LF \\
\hline
\bf{Current out of bounds}  & Quantitative (Integer) $\mathcal{F}_{4}^{Q}$ & The maximum operating current of the backbone is less than or equal to 60\% of the threshold ``out of bounds'', typically established at 90\% of capacity \\
\hline
\bf{Max. temperature} & Quantitative (Real) $\mathcal{F}_{5}^{Q}$  & Maximum registered temperature \\
\hline
\bf{Min. temperature} & Quantitative (Real) $\mathcal{F}_{6}^{Q}$ & Minimum registered temperature \\
\hline
\bf{Delta temperature section} & Quantitative (Real) $\mathcal{F}_{7}^{Q}$  & Difference between the maximum and minimum temperature \\
\hline
\bf{Rain} & Quantitative (Real) $\mathcal{F}_{8}^{Q}$  & Millimeters of rain calculated as the average two hours preceding to the LF \\
\hline
\bf{Cable section} & Quantitative (Real) $\mathcal{F}^{S}$  & Section of the cable, if applicable  \\
\hline
\bf{Backbone electric current} &  Quantitative (Real)  $\mathcal{F}^{EC}$ & Function of electric current that flows in a backbone \\
\hline
\bf{Interruption (breaker)} & TS (Integer) $\mathcal{F}_{1}^{TS}$ & Outages caused by the opening of the \textit{breakers} in the primary station \\
\hline
\bf{Petersen alarms} & TS (Integer) $\mathcal{F}_{2}^{TS}$ &  Alarms detected by the device called ``Petersen's coil'' due to loss of electrical insulation on the power line \\
\hline
\bf{Saving intervention} & TS (Integer) $\mathcal{F}_{3}^{TS}$ & Decisive interventions of the Petersen's coil which have prevented the LF \\
\hline
\end{tabular}
\label{tab:features_description}
\end{table*}

\subsection{Data Description and Preprocessing}
\label{sec:Data_Preprocessing}

Data normalization is a universally important aspect in pattern analysis, which becomes even more crucial when processing patterns characterized by many heterogeneous features.
The numerical data provided by ACEA have been normalized using the affine normalization technique:
\begin{equation}
\label{eq:Affine_normalization}
v=\frac{c-m}{\left (M-m  \right )} \in [0, 1].
\end{equation}

\noindent where $c$ is the original (non-normalized) value; $m$ and $M$ are, respectively, the minimum/maximum values for the specific feature in the considered dataset.

\subsubsection{Temporal Data}
\label{sec:Temporal_Data}
The ``Day start'' and ``Time start'' features (Tab. \ref{tab:features_description}) have been encoded as integer values. The former ranges in $\{0, 1, ..., 364\}$, while the latter in $\{0, 1, ..., 1439\}$, corresponding to the number of minutes in a year.
Normalization of such data follows straightforwardly.

\subsubsection{Spatial Data}
\label{sec:Spatial_Data}

Three types of information regarding the geographical position of a LF are available: the absolute position of the PS where the LF has occurred, and the absolute position of the two SSs delimiting the section of power line where the LF has been detected.
The original coordinates of the geographical position of the LF have been expressed in WGS84 (decimal degrees), the same that it is used in the GPS geolocalization system.
It is reasonable that the information regarding the PSs positions and the absolute locations of the LFs can provide indirectly the information about the amount of electric current flowing in the power line. The main hypotheses that led us to that statement are: (i) the MV lines have a radial distribution with respect to the PSs and their extension is of the order of kilometers, (ii) the portion of power line between the two SSs has an extension of hundreds of meters. 
In addition, the power grid has a meshed structure and it is ``radially'' distributed, so that every MV line is supplied through only one PS.

\begin{figure}[ht!]
 \centering
 \includegraphics[viewport=0 0 564 137,scale=0.55,keepaspectratio=true]{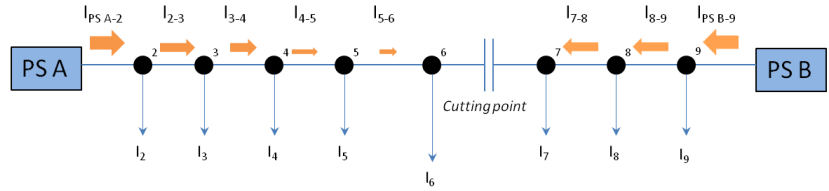}
 \caption{Scheme of a MV backbone.}
 \label{fig:Backbone_Scheme}
\end{figure}

In Fig. \ref{fig:Backbone_Scheme} is depicted a typical scenario: a MV backbone composed by two lines supplied through two distinct PSs; the cutting point is situated in (roughly) the middle point.
From the Kirchhoff's second law, the current $I_{PS\ A-6}$ that the PS denoted by A provides to the MV lines to feed nodes from 2 to 6 is equal to the sum of the currents provided by each substation before the cutting point:
\begin{equation}
\label{eq:Kirchoff}
I_{PS\ A-6}=I_{2}+I_{3}+I_{4}+I_{5}+I_{6}.
\end{equation}

\begin{figure}[ht!]
 \centering
 \includegraphics[viewport=0 0 443 296,scale=0.55,keepaspectratio=true]{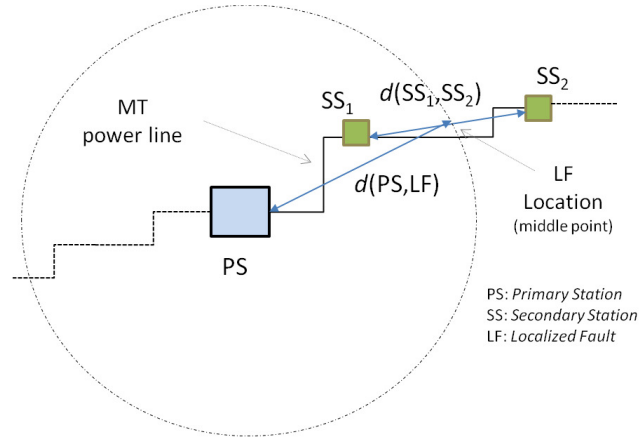}
 \caption{Radial structure of the ACEA grid from the PS and the approximated distance computation (length of red arrow) adopted to reduce the number of available features.}
 \label{fig:Hyp_PS_SS}
\end{figure}

The intensity of the electric current in the MV feeder decreases as we move away from the PS, until the open breaker is reached, given the radial topology of the network, we can roughly evaluate the fault location as the median point between of the line segment connecting the two SSs soon before and soon after the localized fault. Moreover, the distance between the PS location and the fault location can be considered as an indirect measure of the total current provided by the PS in the fault condition. -- see Fig. \ref{fig:Hyp_PS_SS} for a graphical representation.

\begin{figure}[ht!]
 \centering
 \includegraphics[viewport=0 0 919 483,scale=0.35,keepaspectratio=true]{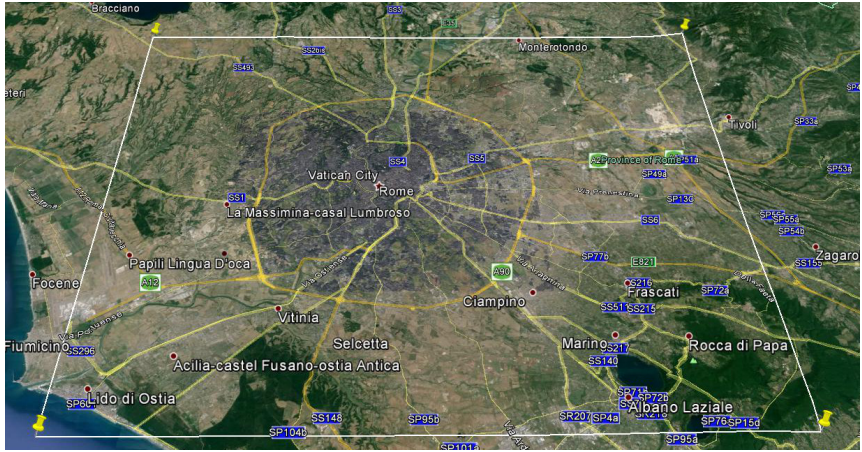}
 \caption{Covered are of Rome and the quadrilateral normalization.}
 \label{fig:Rome_area}
\end{figure}

The maximum spatial resolution of the geographical localization of the LF is therefore defined by the two SS positions.
The distance between two geographical locations is calculated through the Vincenty's algorithm \cite{vincenty_direct_1975}.
The normalization process of the position data is based on the calculation of the largest rectangle including all the PS and SS stations. (see Fig. \ref{fig:Rome_area} for an example).
Hence, applying the affine normalization (\ref{eq:Affine_normalization}), the spatial positions of the LFs are normalized in $[0, 1]$.
The affine normalization is applied also for the distance values among PSs and the positions of the LFs.

\subsubsection{Physical Data}
\label{sec:Physical_Data}

The data describing the physical elements of the power grid is defined by both categorical and quantitative information. 
The normalization of quantitative data (i.e., ``\# Secondary station'', ``Current out of bounds'', and ``Cable section'') is implemented by means of (\ref{eq:Affine_normalization}).

It is well known that a possible cause of faults in distribution systems affecting cables and joints is the abrupt change in the current loads, more than the actual amount of electrical power flowing in the devices. To define a feature taking this effect into account, current measures sampled every 10 minutes have been considered in a main window of 24 hours before the LF occurrence. This time window is divided in two non overlapping sub-windows, $w_{1}$ and $w_{2}$, each of 12 hours. The feature ``Backbone Electric Current'' is finally computed as the absolute difference between the average of current values in each sub-window.
This value is normalized with respect to the possible minimum and maximum values.
Clearly, this single real value is a lossy compression of the information conveyed by the whole TS in a single numerical value. However, we performed this simplification in order to capture the average information about the fluctuations of the values of the electric current observed before the LF.
Future dedicated research works will be focused to the study of the TSs of electric current and their relation/causation with the observed LFs.

\subsubsection{Meteorological Data}
\label{sec:Meteorological_Data}

The meteorological data are acquired by suitable stations located in different areas of Rome. The ``Rain'' feature is calculated as the average millimeters of rain observed in the 2 hours soon before the LF occurrence.

\subsubsection{Short Outages Data}
\label{sec:Time_Series}

Here we describe the data related to the short outages observed before a LF.
With an abuse of notation, we will refer to such sequences of events as time series, although formally such sequences are not sampled with a predefined constant period (those events are registered as they occur).

\begin{figure}[ht!]
 \centering
 \includegraphics[viewport=0 0 387 119,scale=0.7,keepaspectratio=true]{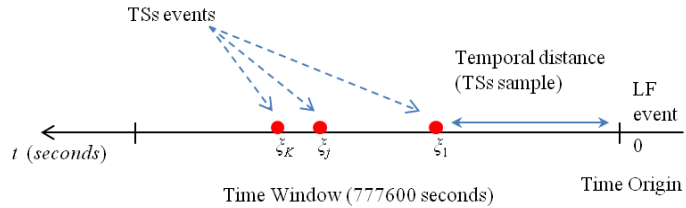}
 \caption{Representation of TSs of outages happened before a LF.}
 \label{fig:Time_Series_rapres}
\end{figure}

We consider three types of events that can be associated to the ``short outages'' type (see Sec. \ref{sec:SG}). The considered TS of events are: ``Interruption (breaker)'', the ``Petersen alarms'', and the ``Saving intervention'' (see Tab. \ref{tab:features_description} for details).
The short outages events are represented as variable-length sequences, which contain the temporal distances (expressed in seconds) from the subsequent LF (see Fig. \ref{fig:Time_Series_rapres} for a graphical example).
The time window in which those events fall spans across three months (i.e., we search in the three months preceding a LF).
A TS $S^{i}$ of $K_{i}$ outage events is defined as follows:
\begin{equation}
\label{eq:TS_set}
S^{i}=\left [ \xi _{1}^{i},\xi _{2}^{i}, ..., \xi _{K_{i}}^{i} \right ],
\end{equation}
where $\xi$ is the temporal distance from the LF (considered as the origin), $i\in \{1,2,3\}$ is the index distinguishing the three aforementioned types of outages, and $K_{i}$ is the number of events for the $i$-th type of outage.

Normalization of such data is performed as follows. Given a dissimilarity measure for TSs (see Sec. \ref{sec:Time_Series_Data}), we pre-compute, for each type of short outage feature, the dissimilarity matrix, i.e.  a symmetric square matrix containing all the possible pairwise dissimilarity values between patterns in the considered dataset. Successively, we apply the normalization dividing the dissimilarity value between two TSs with the maximum value in the dissimilarity matrix.

\subsection{The Proposed One-class Classifier}

As a consequence of the difficulty of modeling useful (and meaningful) instances of non-faults in the considered SG, we designed a OCC for the purpose of recognizing LFs.
Such a goal is implemented by building a OCC relying on clustering techniques. The idea of using clusters for modeling a region of the ``fault representation space'', $\mathcal{F}$, containing target patterns representing LFs, is reasonable and also intuitive. The underlying assumption is that, similar statuses of the SG have similar chances of generating a LF, assumption that is reflected by the cluster model.

A dataset of FPs is partitioned in $k$ (disjoint) clusters, where each cluster contains faults having similar features.
Accordingly, the most important component of the OCC system is the core dissimilarity measure $d: \mathcal{F}\times\mathcal{F}\rightarrow\mathbb{R}^+$, which assigns a dissimilarity value to a pair of FPs.
The partition, as well as other parameters that will be described in the following, constitute the model of the OCC.

We would like to remark the need of designing an ad-hoc OCC. The application that we face in this study deals with highly structured data, which are processed by means of a weighted aggregation of several heterogeneous dissimilarity measures. This results in a FP representation space $(\mathcal{F}$, $d(\cdot, \cdot))$ that is not Euclidean (nor it is metric). Therefore, using more consolidated methods, such as the SVDD cited in Sec. \ref{sec:occ_prob}, is not straightforward and it would require the definition of particular (positive definite) kernel functions tailored to the problem at hand.

\subsubsection{The Dissimilarity Measure Among FPs}
\label{sec:dissimilarity_measure}

Let $\mathcal{S}\subset\mathcal{F}$ be a dataset of FPs. A FP $x\in\mathcal{S}$ is described as
\begin{equation}
\label{eq:data_obj}
x=\left \{ \mathbf{F}_{1},\mathbf{F}_{2}, ...,\mathbf{F}_{m} \right \},
\end{equation}
where the $l$-th feature, $\mathbf{F}_{l}$, $1 \leq  l \leq m$, lies in its specific feature space $\mathcal{F}_{l}$. Hence, each pattern $x$ lies on the $m$-fold product feature space $\mathcal{F}=\mathcal{F}_{1}\times \mathcal{F}_{2}\times ...\times \mathcal{F}_{m}$.

Given two FPs $x, y \in \mathcal{S}$, the proposed weighted dissimilarity measure reads as:
\begin{equation}
\label{eq:Custom_Distance}
d(x, y; \underline{\mathbf{w}})= \sqrt{\sum_{j=1}^{m} w_{j}\times \left(x_{j} \circleddash y_{j} \right)^2},
\end{equation}
where the $\circleddash$ operator represents a generic dissimilarity measure, and $w_j\in[0, 1]$ is the weight related to the $j$-th feature. In practice, Eq. \ref{eq:Custom_Distance} computes the weighted $l_2$ norm of the vector containing the dissimilarity values calculated feature-wise. However, since not all dissimilarity measures implementing the $\circleddash$ operator are metrics, the resulting dissimilarity measure (\ref{eq:Custom_Distance}) is not metric and hence it does not induce a metric space. This aspect is carefully taken in consideration into the design of our OCC system.
The weights $\underline{\mathbf{w}}\in[0, 1]^m$ are suitably optimized during the training phase of the OCC by means of a Genetic Algorithm (GA). In other words, the overall system is in charge to synthesize a classification model by a clustering technique, learning the parameters of the dissimilarity measure yielding the most appropriate clusters able to define the faults decision regions. For this reason the proposed OCC system fully belongs to the Metric Learning framework \cite{Metric_learning}.

In the following paragraphs, we describe the implementations of $\circleddash$, i.e., the specific dissimilarity measures tailored for each specific FP component.
The nature of the feature, $\mathcal{F}_{i}$, will be denoted using the same notation of Tab. \ref{tab:features_description}.

\paragraph{Categorical Data}
\label{sec:Categorical_Data}

Categorical attributes, also referred to as nominal attributes, are data without an ``meaningful'' ordering (see Tab. \ref{tab:features_description} for the data treated as nominal).
Let $\mathcal{F}^{c}=\left \{ \eta_{1} ,\eta_{2},..., \eta_{n} \right \}$ be the set of all categorical features of the entire dataset, each described by $d$ categorical attributes: $\nu_{1} ,\nu_{2},..., \nu_{d}$. Let us define the domain of the attribute $\nu_{j}$, $\mathrm{DOM}( \nu_{j} )=  \left \{ A_{j_1},A_{j_2},...,A_{j_{n(j)}} \right \}$, where $ A_{j_l}$  $(1\leqslant l \leqslant n(j) )$ is the set of possible values for the categorical attribute $\nu_{j}$, and $n(j)$ is its cardinality. We consider the well-known simple matching distance, defined as follows:
\begin{equation}
\label{eq:SimpleMatch_Distance}
\delta(x,y)=\left\{\begin{matrix}
0 & x=y, \\ 
1 & x\neq y.
\end{matrix}\right.
\end{equation}

Let $x^{c}$ and $y^{c}$ be the projections on the categorical feature space $\mathcal{F}^{c}$ of two generic patterns $x, y$. The dissimilarity measure between the two categorical objects described by $d$ categorical attributes is implemented as:
\begin{equation}
\label{eq:SimpleMatch_Distance_complete}
d^{c}(x^{c}, y^{c})=\frac{1}{d}\sum_{j=1}^{d}\delta (x_{j}^{c},y_{j}^{c}).
\end{equation}

\paragraph{Quantitative Data}
\label{sec:Quantitative_Data}
As concerns the quantitative data (see Tab. \ref{tab:features_description}) we distinguish between (i) ``Normal'' quantitative data and (ii) ``Special'' quantitative data. The former type includes both numerical and integer values (normalized in $[0, 1]$); the operator $\circleddash$ is implemented by the absolute difference: $d^{N} =\left | x-y\right |$.

For integer values describing information related to timestamps temporal information, such as the day in which the LF happened and the time of day, we defined a particular dissimilarity measure implementing $\circleddash$, called \textit{circular difference}.
Given an ordered set of integer numbers $\{0, 1, ..., a\}$, the circular difference among any $x,y$ in this set is given by
\begin{equation}
\label{eq:Circular_diff}
d^{CD}(x, y; a)=\min(\left |x-y  \right |,a-\left |x-y  \right |),
\end{equation}
where $a$ is considered as a parameter.
For ``Day start'' and ``Time start'' the maximum value for $a$ in (\ref{eq:Circular_diff}) is 364 and 1439, respectively.
The implementation of the circular difference is designed to avoid that pairs of close days or timestamps give raise to high dissimilarity values.

``Special'' quantitative data are normalized in the range $[0, 1]$, but can assume also a special symbol, $\epsilon$, indicating the ``not applicable'' condition.
It is the case for the ``Cable section'' feature, since for LFs not related to cables this field is undefined. The dissimilarity measure $d^{S}: \{[0, 1] \cup \epsilon\} \times \{[0, 1] \cup \epsilon\} \rightarrow [0, 1]$ for two special quantitative values $x,y \in \mathcal{F}^{S}$ is defined as follows:
\begin{equation}
\label{eq:Special_diff}
d^{S}=\left\{\begin{matrix}
|x-y|  & x \neq \epsilon \wedge  y \neq \epsilon, \\ 
  1    &  x =\epsilon \vee y=\epsilon, \\ 
  0    & x=\epsilon \wedge  y=\epsilon.
\end{matrix}\right.
\end{equation}

\paragraph{Time Series Data}
\label{sec:Time_Series_Data}

The Dynamic Time Warping (DTW) is a well-known algorithm to find an optimal alignment between two sequences of objects (i. e. Time Series) of variable length.
The use of DTW as dissimilarity measure for sequences of generic objects is well-established in many applications, such as biology, finance, multimedia, and image analysis \cite{PiyushShanker20071407,t2vsdiss__ifsanafips2013}.
An in-depth description of DTW algorithm can be found in \cite{muller_dtw}.

Following the notation introduced in Sec. \ref{sec:Time_Series}, the dataset consists in three types of TSs, $S^{i}$, with  $i\in\{1,2,3\}$. Each one represents a vector belonging to the TSs feature vector subspace, $\mathcal{F}_{i}^{TS}, i\in\{1, 2,3\}$. 
Let $x, y \in \mathcal{F}_{i}^{TS}$ be two TSs, the dissimilarity measure between them is computed as as a function $d^{TS}: \mathbb{R}^{m}\times \mathbb{R}^{n}\rightarrow \mathbb{R}$:
\begin{equation}
\label{eq:DTW}
d^{TS}(x, y)=\mathrm{DTW}(x,y),
\end{equation}
where $m$ and $n$ are the lengths of $x$ and $y$, respectively.

\subsubsection{Model Definition and the Classifier Decision Rule}
\label{sec:model}

\begin{figure}[ht!]
 \centering
 \includegraphics[viewport=0 0 502 200,scale=0.6,keepaspectratio=true]{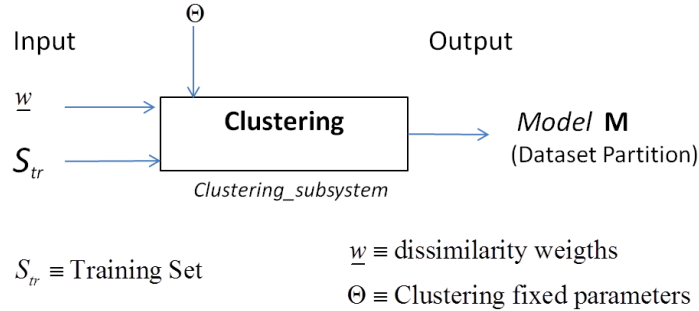}
 \caption{The OCC model is defined by a partition of the training set $\mathcal{S}_{tr}$.}
 \label{fig:Clustering_model}
\end{figure}

The most important component of the OCC model is the partition $P$ (i.e., a set of clusters), determined on the training set $\mathcal{S}_{tr}$. The partition is obtained through a clustering algorithm -- see Fig. \ref{fig:Clustering_model} for an overview.

A hard partition of order $k$ is a collection of $k$ disjoint and non-empty clusters, $P=\{\mathcal{C}_{1}, \mathcal{C}_2, ..., \mathcal{C}_k\}$.
Each cluster $\mathcal{C}_i\in P$ is synthetically described by a \textit{representative} element, which we denote as $c_i=R(\mathcal{C}_i)$; let $R(P)=\{c_1, c_2, ..., c_k\}$ be the set of representatives of the partition $P$.
The representative of $\mathcal{C}_i$ is computed as the element $c_i$ that minimizes the sum of distances (MinSOD) \cite{delvescovo+livi+rizzi+frattalemascioli2011}:
\begin{equation}
\label{eq:minsod}
c_i = \argmin_{x_j\in\mathcal{C}_i} \sum_{x_k\in\mathcal{C}_i} d(x_j, x_k).
\end{equation}

A cluster representative $c_i$ is, in a sense, the prototype of a \textit{typical fault scenario} individuated in $\mathcal{S}_{tr}$.
As a consequence, the information provided by the cluster $\mathcal{C}_i$ as a whole is useful to conceive a region of the pattern space ``around'' $c_i$, which describes similar fault scenarios.
By defining $\delta(\mathcal{C}_i)\geq0$ as a measure of cluster extent, we can construct the decision region associated to each cluster $\mathcal{C}_j$, used to implement the classification rule.
The cluster extent can be computed as the average/maximum intra-cluster dissimilarity value or by considering their standard deviation, for instance.
In the case of the average, the expression reads as:
\begin{equation}
\delta(\mathcal{C}_i) = \frac{1}{|\mathcal{C}_i|-1}\sum_{x_k\in\mathcal{C}_i} d(c_i, x_k)).
\end{equation}

In addition to $\delta(\mathcal{C}_i)$, we consider also a tolerance parameter, $\sigma_i\geq0$, for defining the decision region.
The decision region derived from a cluster $\mathcal{C}_i$ is hence defined by the quantity $B(\mathcal{C}_i)=\delta(\mathcal{C}_i)+\sigma_i$, which actually defines the neighborhood of $c_i$.

\begin{figure}[ht!]
 \centering
 \includegraphics[viewport=0 0 631 224,scale=0.5,keepaspectratio=true]{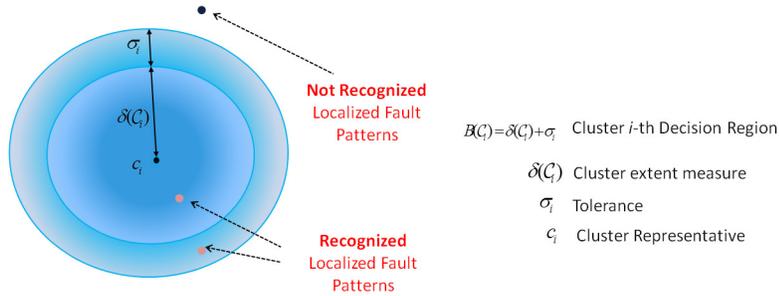}
 \caption{The cluster decision region and its characterizing parameters.}
 \label{fig:Region_model}
\end{figure}

Fig. \ref{fig:Region_model} provides a schematic overview of a cluster model and its use in the process of classifying a test pattern $\bar{x}$.
The classification rule for a test pattern $\bar{x}$ operates in two stages. First, the nearest cluster representative $c^*\in R(P)$ is individuated according to the following expression:
\begin{equation}
\label{eq:1}
c^* = \mathop {\arg \min }\limits_{{c_j} \in R(P)} d(\bar x, {c_j}).
\end{equation}

The second step consists in comparing the dissimilarity value $d(\bar{x}, c^*)$ with $B(\mathcal{C}^*)$.
We define a binary-valued function $h(\cdot)$ that performs the hard classification:
\begin{equation}
\label{eq:class_rule}
h(x)=
\begin{cases}
1 & \textrm{if}\ d(\bar x, c^*) \le B(\mathcal{C}^*),  \\ 
0 & \textrm{otherwise}.
\end{cases}
\end{equation}

Along with the hard classification (\ref{eq:class_rule}), we developed a mechanism based on fuzzy sets to provide the user with a measure of ``reliability'' associated to the decisions.
This topic is discussed in the following dedicated section.

\subsubsection{Evaluating the Reliability of the Classification}
\label{sec:class_reliability}

\begin{figure}[ht!]
 \centering
 \includegraphics[viewport=0 0 728 319,scale=0.4,keepaspectratio=true]{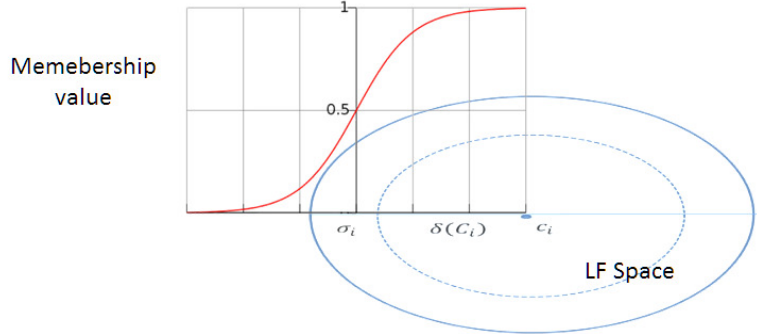}
   \caption{Sigmoidal membership function associated to the decision region.}
    \label{fig:Reliability}
\end{figure}

A Boolean decision regarding if a new test pattern (i.e., a given SG status) is a fault or not, is operatively reasonable. However, it is important to provide the user also with an additional measure that quantifies the reliability of such a decision. This becomes even more appropriate in the particular OCC setting.
For this purpose, we equip each cluster $\mathcal{C}_i$ with a suitable membership function, denoted in the following as $\mu_{\mathcal{C}_i}(\cdot)$.
In practice, we generate a fuzzy set over $\mathcal{C}_i$.
The membership function allows us to quantify the uncertainty (expressed by the membership degree in $[0, 1]$) of a decision about the recognition of a test pattern.
Fig. \ref{fig:Reliability} depicts this idea by an intuitive illustration.
Membership values close to either 0 or 1 denote ``certain'' and hence reliable decisions. When the membership degree assigned to a test pattern is close to 0.5, there is no clear distinction about the fact that such a test pattern is really a fault or not (regardless of the correctness of the Boolean decision).

For this purpose, we used a parametric sigmoid model for $\mu_{\mathcal{C}_i}(\cdot)$, which is defined as follows:
\begin{equation}
\label{eq:membership}
\mu_{\mathcal{C}_i}(x) = \frac{1}{1+\exp((d(c_i, x)-b_i)/a_i)},
\end{equation}
where $a_i,b_i\geq 0$ are two parameters specific to $\mathcal{C}_i$, and $d(\cdot, \cdot)$ is the dissimilarity measure (\ref{eq:Custom_Distance}). Notably, $a_i$ is used to control the steepness of the sigmoid (the lower the value, the faster the rate of change), and $b_i$ is used to translate the function in the input domain.
If a cluster (that models a typical fault situation found in the training set) is very compact, then it describes a very specific fault scenario. Therefore, no significant variations should be accepted to consider test patterns as members of this cluster.
The converse is also true. If a cluster is characterized by a wide extent, then we might be more tolerant in the evaluation of the membership.
Accordingly, the parameter $a_i$ is set equal to $\delta(\mathcal{C}_i)$.
On the other hand, we define $b_i = \delta(\mathcal{C}_i) + \sigma_i/2$. This allows us to position the part of the sigmoid that changes faster right in-between the area of the decision region determined by the dissimilarity values falling in $[B(\mathcal{C}_i)-\sigma_i, B(\mathcal{C}_i)]$.

Finally, the soft decision function, $s(\cdot)$, is defined as
\begin{equation}
\label{eq:soft_decision}
s(\bar x) = \mu_{\mathcal{C}^*}(\bar x),
\end{equation}
where $\mathcal{C}^*$ is the cluster satisfying Eq. \ref{eq:1}.

The evaluation of Eq. \ref{eq:soft_decision} over a test set $\mathcal{S}_{ts}, n=|\mathcal{S}_{ts}|$, yields $n$ membership degrees, each assigned to a specific pattern of $\mathcal{S}_{ts}$.
We can evaluate the overall reliability of the decisions taken on $\mathcal{S}_{ts}$ by considering a fuzzy set, say $\mathcal{M}$, characterized by those $n$ membership degrees.
We can implement such an evaluation measure by calculating the fuzzy entropy \cite{Livi_ga_2013} of $\mathcal{M}$. A possible expression for the fuzzy entropy reads as
\begin{equation}
\label{eq:fe}
\frac{\mathrm{card}(\mathcal{M}\cap \mathcal{M}^c)}{\mathrm{card}(\mathcal{M}\cup \mathcal{M}^c)}\in[0, 1],
\end{equation}
where $\cap$ and $\cup$ are defined as the minimum and maximum operators, respectively, and the cardinality of the resulting fuzzy set is taken as the sum of the membership degrees.
Fuzzy entropy values close to zero would indicate that, overall, the decisions are reliable. Conversely, if the fuzzy entropy is close to one then the decisions are highly unreliable.

It is worth stressing that the reliability measures herein discussed should not be confused with the measures of correctness of the recognition (i.e., the evaluation of the correctness of the discrimination among target and non-target patterns).

\subsubsection{Training of the OCC by the \textit{k}-means Algorithm}
\label{sec:kmeans_training}

We propose a learning strategy to synthesize the OCC model that is based on the well-known \textit{k}-means \cite{Jain:2010:DCY:1755267.1755654}. Since our feature space is non-metric and we represent each cluster by the MinSOD pattern, in the technical literature this algorithm is usually referred to $k$-medoids \cite{Park20093336}. This clustering procedure depends on an integer parameter, $k$, defining a priori the partition order.
The dissimilarity measure described in Sec. \ref{sec:dissimilarity_measure} depends on a vector of weights, $\underline{\mathbf{w}}$. Moreover, the decision regions -- Sec. \ref{sec:model} -- are defined by the thresholds $\sigma_i$.
Setting those parameters, denoted $p_j=[\underline{\mathbf{w}}_j, \underline{\sigma}_j]$, is of utmost importance, and of course it has a significative influence on the results yielded by \textit{k}-means.

For this reason, a GA is employed to find the best-performing values for $p_j$, i.e., those maximizing the following objective function:
\begin{equation}
\label{eq:obj_kmeans}
f(p_j) = \alpha A(\mathcal{S}_{vs}) + (1-\alpha) \sum_{i=1}^{k} 1-\sigma_i.
\end{equation}

\begin{figure}[ht!]
 \centering
 \includegraphics[viewport=0 0 704 330,scale=0.5,keepaspectratio=true]{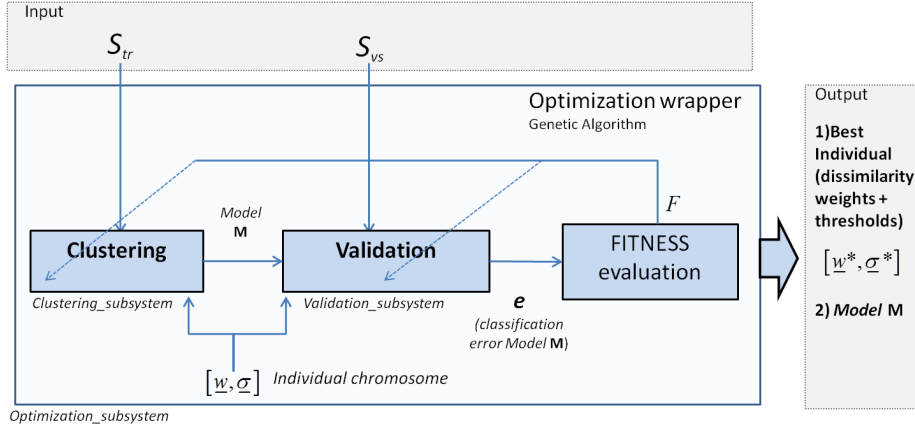}
 \caption{Block diagram depicting the classification model synthesis procedure.}
 \label{fig:System_model}
\end{figure}

In (\ref{eq:obj_kmeans}), $A(\mathcal{S}_{vs})$ is the performance measure achieved on $\mathcal{S}_{vs}$ (we specify the nature of such a measure in the experiments section).
The second term in (\ref{eq:obj_kmeans}) defines a constrain for the (average) cluster extent. The GA is in charge to find the parameters, $p_j$, that minimize the $l_1$ norm of the tolerances used to define the decision regions, while at the same time providing an effective performance in terms of recognition.
Fig. \ref{fig:System_model} shows a diagram illustrating the optimization stage as a whole.

As concerns the learning phase, it is well-known that the \textit{k}-means algorithm is sensitive to the adopted cluster initialization strategy; here we used a fast randomized initialization.
To compensate this fact, the current version of the OCC takes as external parameter the $k$ value and a classification model is synthesized for each $k$ in a given user-defined range $k_{\rm{min}}, k_{\rm{max}}$.
For each $k$ in this range, we synthesize three models with different random initializations. The fitness (\ref{eq:obj_kmeans}) associated to a candidate solution $p_j$ is hence the average of the fitness calculated for those three models.
In the test phase, we use a majority voting scheme to decide if a given test pattern falls in one of the synthesized decision regions or not.

\section{Experimental Evaluation}
\label{sec:experiments}

In Sec. \ref{sec:exp_setting} we introduce the main experimental setting adopted in this paper.
Successively, we discuss the obtained results in terms of quality of the recognition, respectively on synthetically generated data (Sec. \ref{sec:synth}), on some well-known UCI datasets (Sec. \ref{sec:comparison}), and on the data provided by ACEA (Sec. \ref{sec:acea}).
Although the goal of this paper is to present and discuss the results on the ACEA data, we performed experiments also on some UCI datasets to offer a comparison with respect to well-known OCCs on a more established groundwork.

\subsection{Experimental setting}
\label{sec:exp_setting}

A (one-class) classification problem instance is defined as a triple of disjoint sets, namely training set ($\mathcal{S}_{tr}$), validation set ($\mathcal{S}_{vs}$), and test set ($\mathcal{S}_{ts}$).
Given a specific parameters setting, a classification model is synthesized on $\mathcal{S}_{tr}$ and it is validated on $\mathcal{S}_{vs}$. The generalization capability of the optimized model is computed on $\mathcal{S}_{ts}$.

The proposed OCC produces both hard (\ref{eq:class_rule}) and soft decisions (\ref{eq:soft_decision}) on each test pattern.
In the hard decision case, we evaluate the recognition performance of the classifier by exploiting the confusion matrix.
In particular, we consider the false positive rate (FPR), recall, precision, and accuracy \cite{Fawcett:2006:IRA:1159473.1159475}.
On the other hand, in the soft decision case we quantify the correctness of the classifier by computing the area under the ROC curve (AUC) \cite{Fawcett:2006:IRA:1159473.1159475} generated by interpreting the membership degrees (\ref{eq:soft_decision}) as suitable ``scores'' assigned by the classifier to the test patterns.

The OCC parameters defining the model are optimized by means of a GA, which is guided by the objective shown in Eq. \ref{eq:obj_kmeans}; $\alpha=0.8$ has been used as a suitable setting for the problem at hand. 
In (\ref{eq:obj_kmeans}), we implement the accuracy term as the accuracy elaborated from the confusion matrix.
In the voting scheme used during the cross-validation, it is possible to obtain more than one model scoring the same highest value of accuracy on $\mathcal{S}_{vs}$. In such a case, we chose the model characterized by the soft decisions denoting the lowest fuzzy entropy (\ref{eq:fe}).
This would help in choosing a model that able to correctly discriminate fault patterns, maximizing its reliability.
The adopted GA performs stochastic uniform selection, Gaussian mutation, and scattered crossover (with crossover fraction of 0.8).
It implements a form of elitism that imports the two fittest individuals in the next generation; the population size is is kept constant throughout the generations and equal to 50 individuals.
The stop criterion is defined by considering a maximum number of iterations (250) and checking the variations of the best individual fitness.

\subsection{Tests on synthetic data}
\label{sec:synth}

\begin{figure}[ht!]
 \centering
 \includegraphics[viewport=0 0 407 344,scale=0.5,keepaspectratio=true]{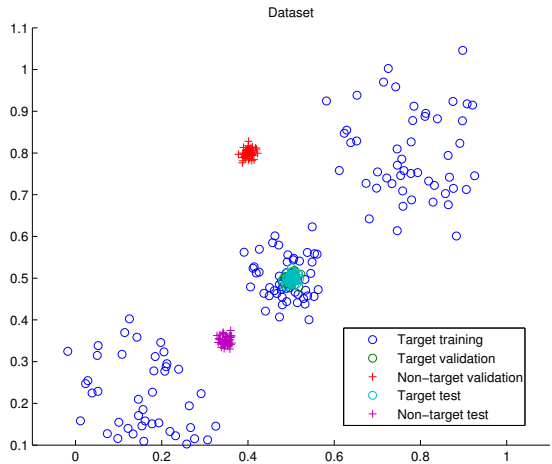}
 \caption{Synthetic problem.}
 \label{fig:test1}
\end{figure}

\begin{figure}[ht!]
\centering

\subfigure[First model. Fuzzy entropy is 0.0007.]{
\includegraphics[viewport=0 0 365 284,scale=0.45,keepaspectratio=true]{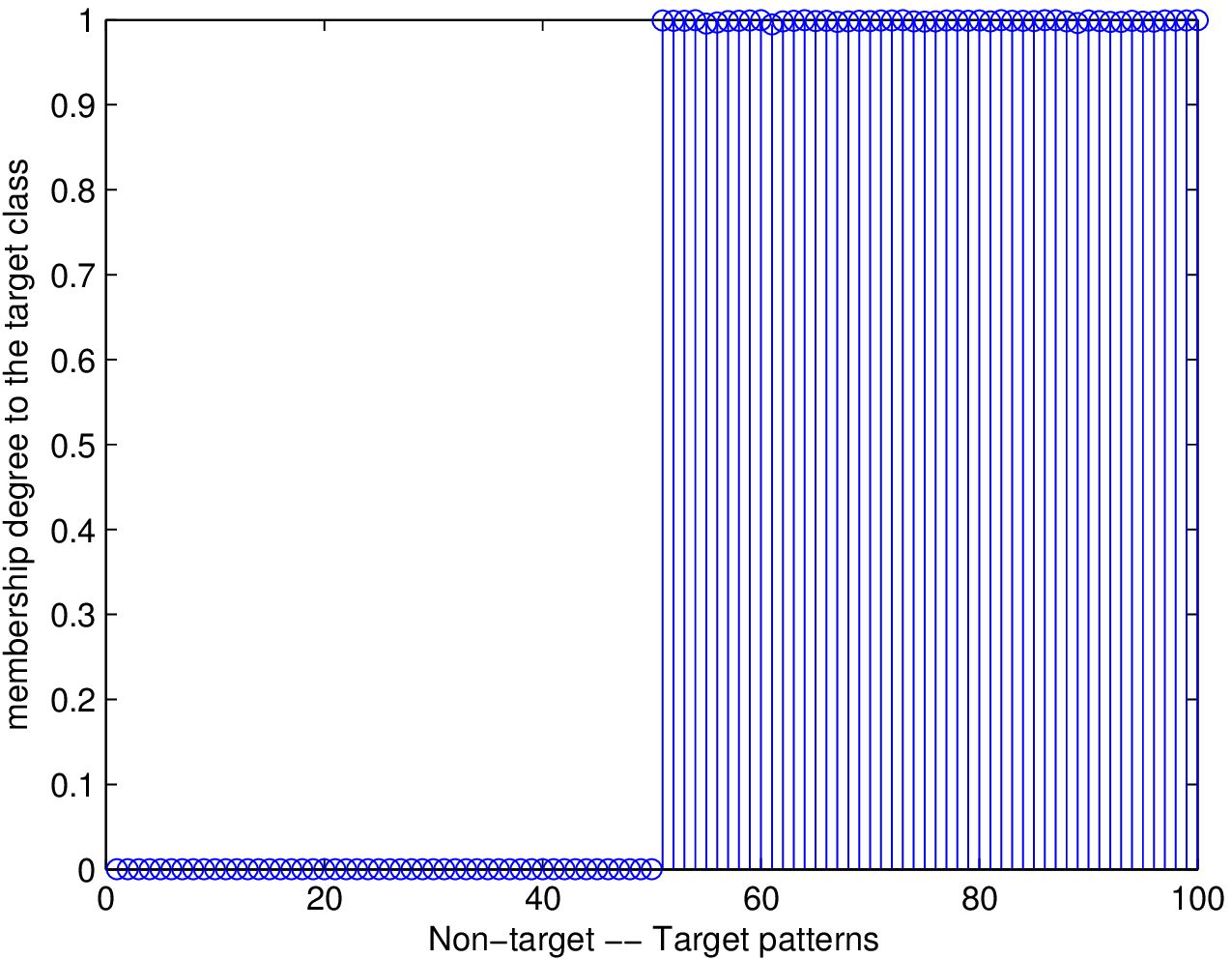}
\label{fig:test1_k3_m1}}
\quad
\subfigure[Second model. Fuzzy entropy is 0.0599.]{
\includegraphics[viewport=0 0 365 284,scale=0.45,keepaspectratio=true]{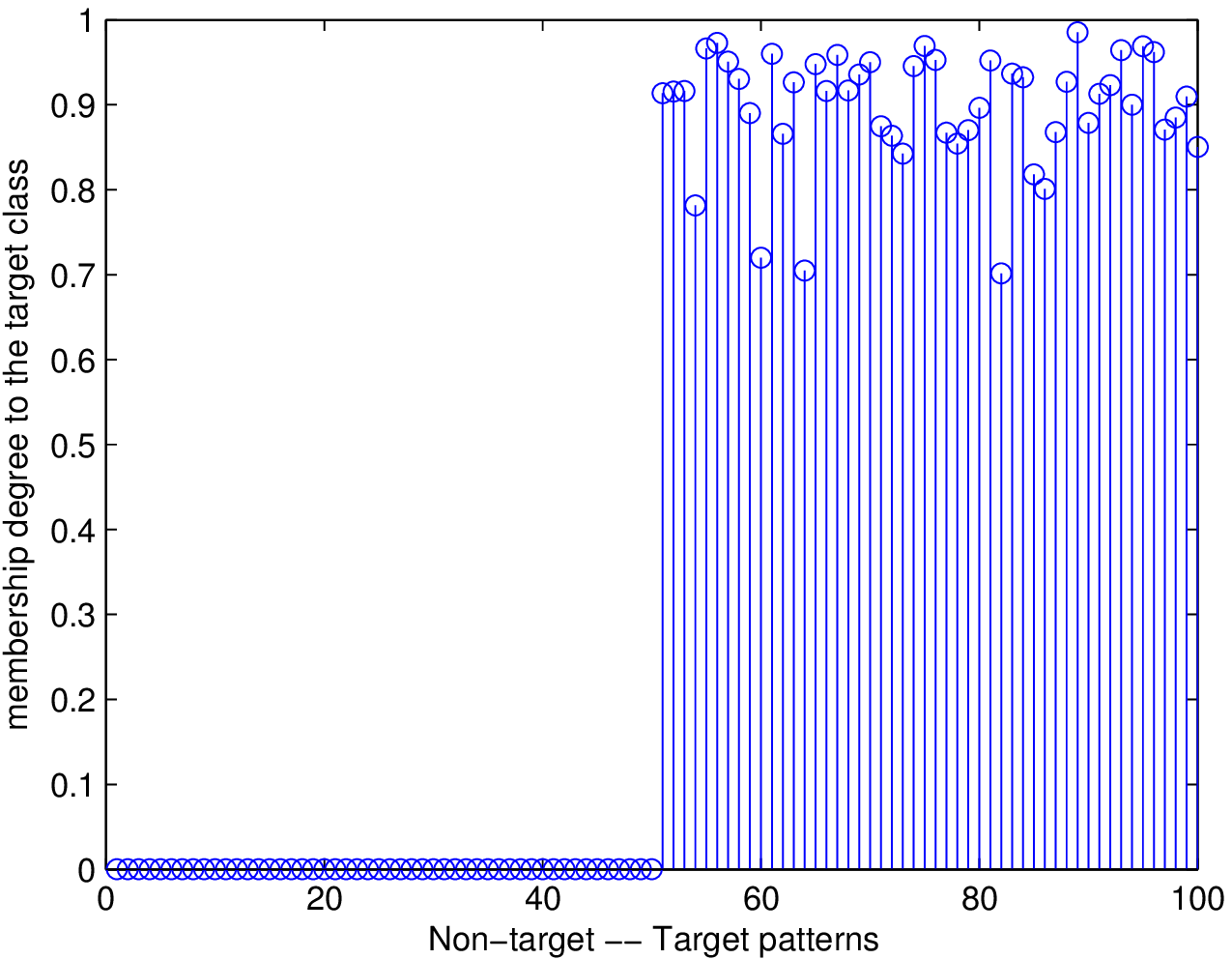}
\label{fig:test1_k3_m2}}

\subfigure[Third model. Fuzzy entropy is 0.0518.]{
\includegraphics[viewport=0 0 365 284,scale=0.45,keepaspectratio=true]{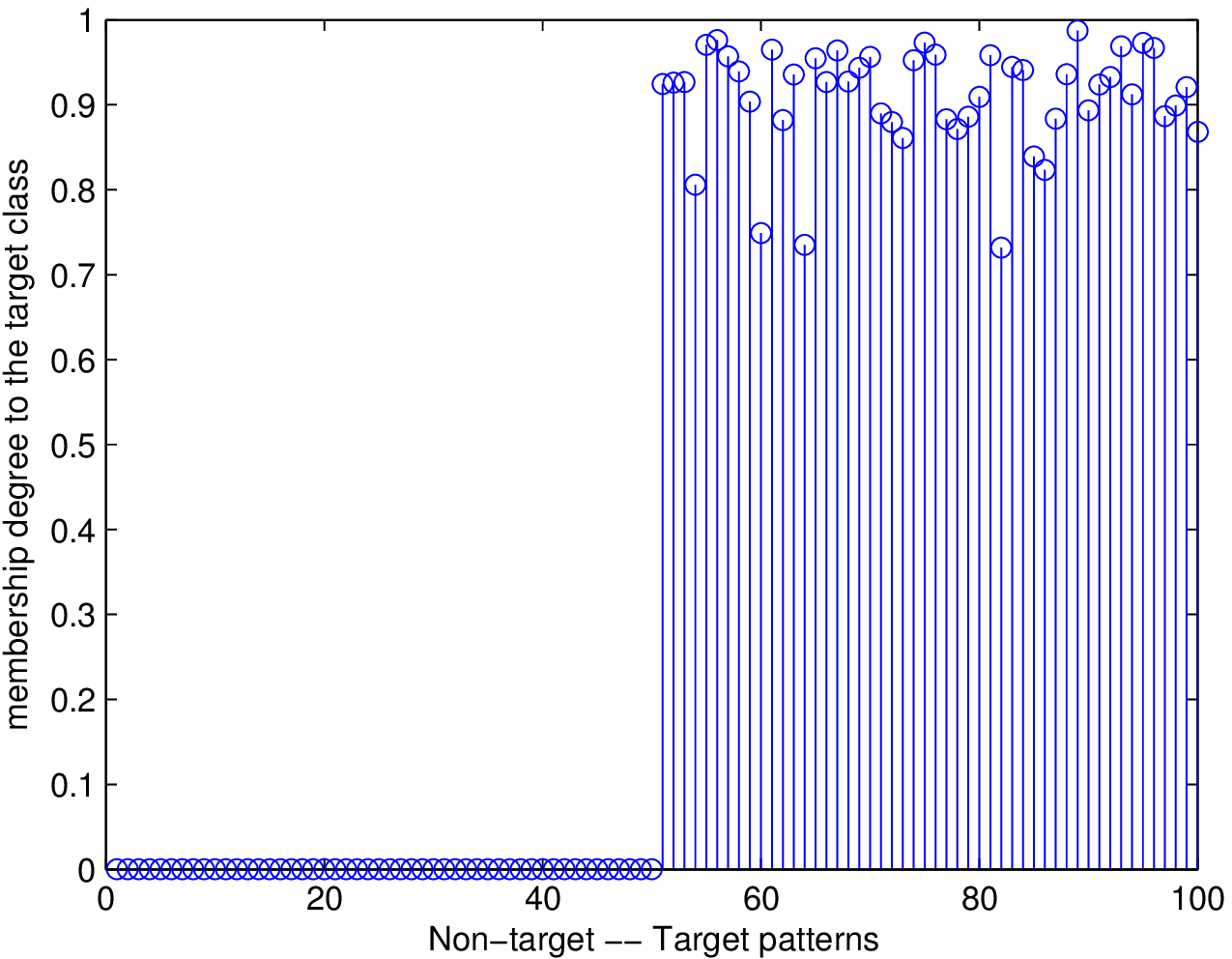}
\label{fig:test1_k3_m3}}

\caption{Results of the soft decisions obtained by the three models synthesized with the \textit{k}-means using the best-performing solution.}
\label{fig:test1_results}
\end{figure}

Fig. \ref{fig:test1} shows the first synthetic test that we have conceived just to show the functioning of the proposed OCC.
The target patterns used for training the OCC are distributed in three Gaussian shaped, well-separated, clusters.
Patterns used for testing are clearly highly recognizable.
The \textit{k}-means is executed with $k=3$ and by synthesizing three different models to be used for the majority voting mechanism.
The accuracy obtained on the test set with the \textit{k}-means is equal to one for all three models.
In Fig. \ref{fig:test1_results}, we report the soft decisions on the test set of the three best-performing models. While the hard decisions are all correct, we note that with the first model we obtain more reliable decisions. In fact, the computed fuzzy entropy is much lower than the other two (the fuzzy entropy is almost zero). As a consequence, the OCC chooses the results of this model as the final output.

\begin{figure}[ht!]
\centering

\subfigure[Dataset distribution.]{
\includegraphics[viewport=0 0 411 378,scale=0.45,keepaspectratio=true]{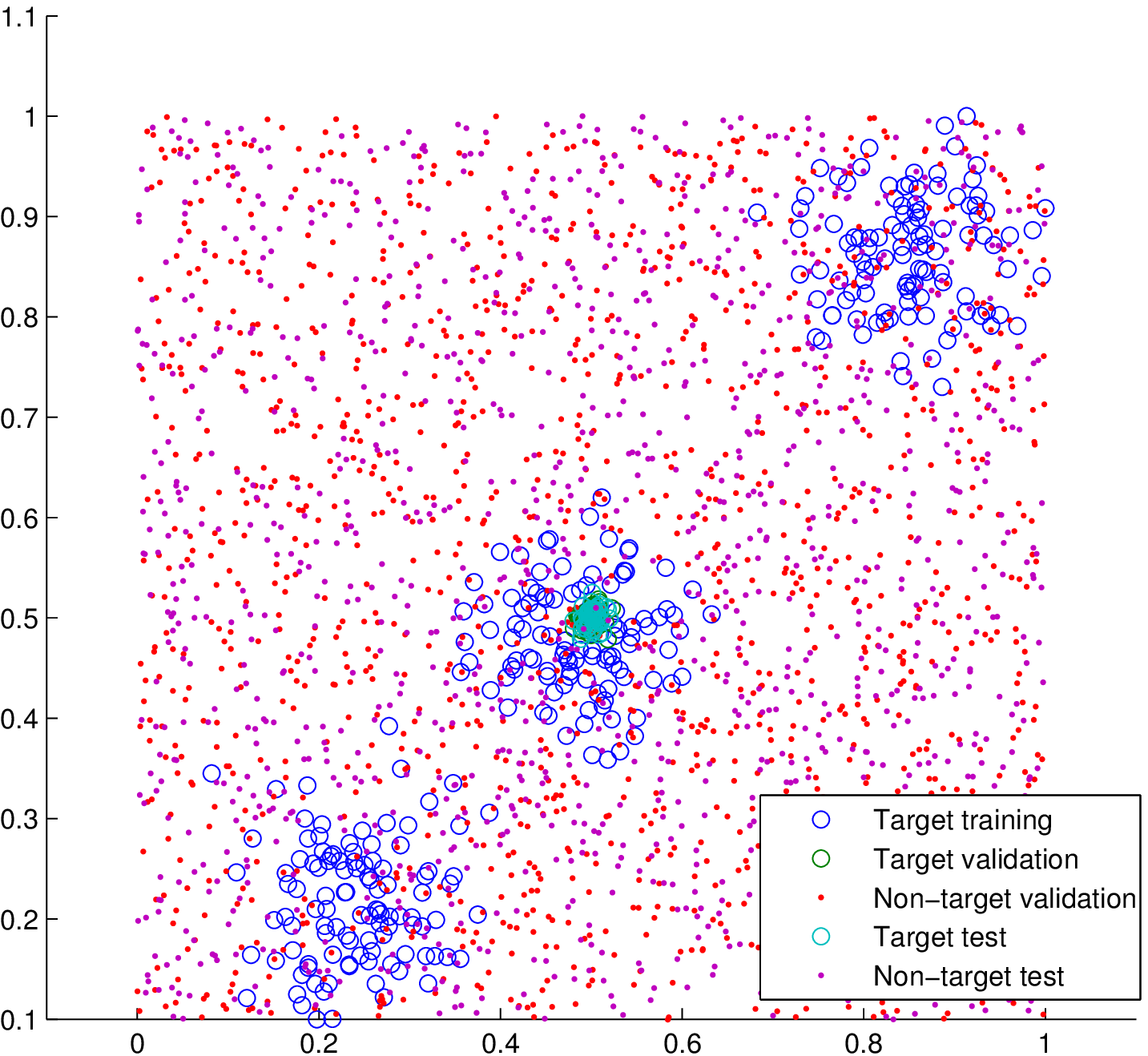}
\label{fig:test2}}
~
\subfigure[Fuzzy entropy of the best model is 0.0352.]{
\includegraphics[viewport=0 0 417 323,scale=0.45,keepaspectratio=true]{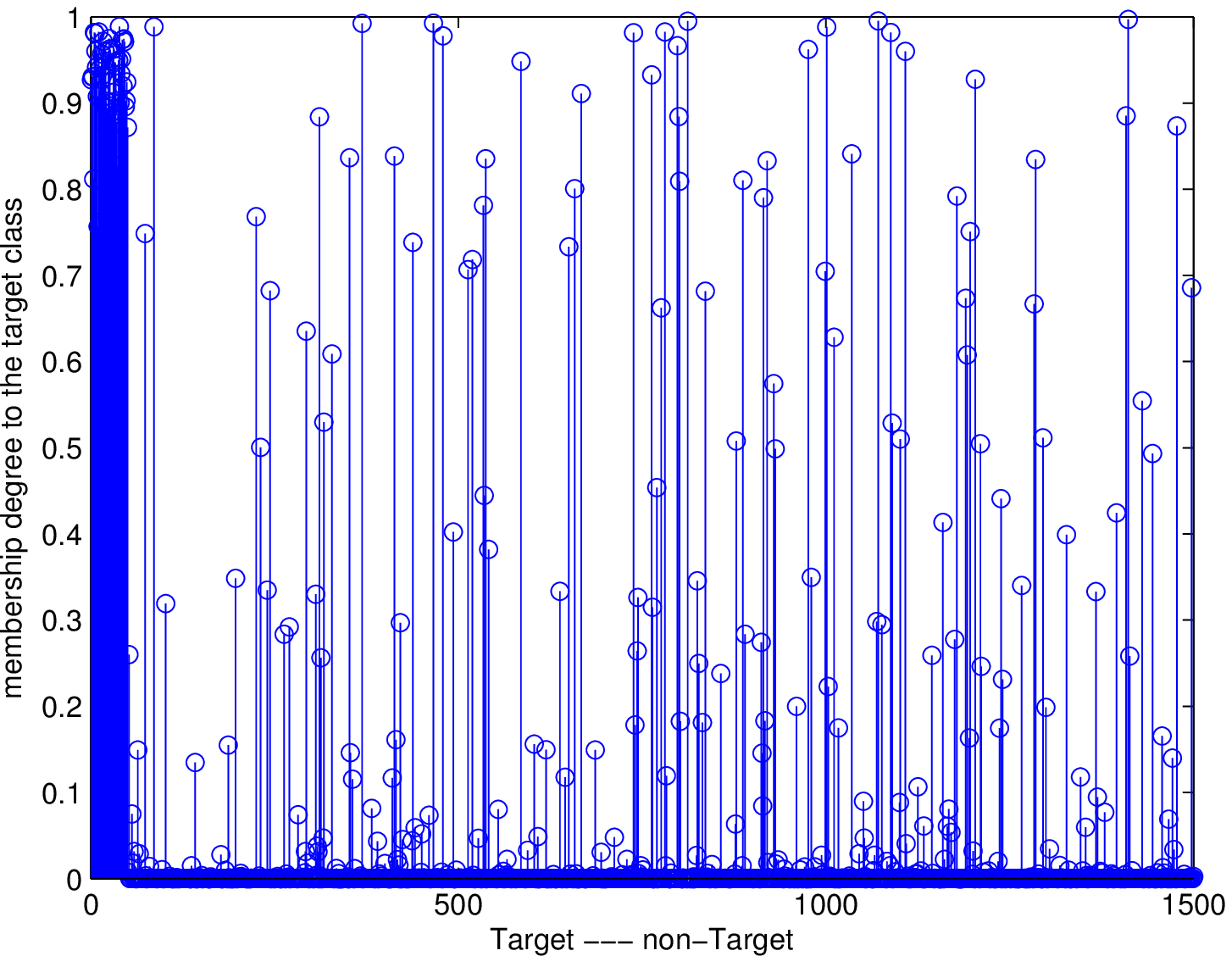}
\label{fig:test2_mf}}

\caption{Dataset distribution and related fuzzy membership values calculated by the best model.}
\label{fig:test2_results}
\end{figure}

We now move to a synthetic test designed to provide a justification for the generation of the non-fault (in the following, non-target and non-fault will be treated as synonyms) patterns that we used within the ACEA data (discussed later in Sec. \ref{sec:acea}).
Fig. \ref{fig:test2} illustrates the considered setting. Training, validation, and test target patterns are still grouped in three well-separated Gaussian clusters. Non-target patterns used for validating and testing a model are distributed uniformly over the $[0, 1]^2$ domain; those patterns outnumber the target patterns, since some non-target patterns will fall in the fault decision region.
By testing the OCC over such data, we expect to observe an ``implicit'' FPR that is proportional to the number of (training set) target patterns. In other terms, we expect to observe $FPR\simeq |\mathcal{S}_{tr}|/|\mathcal{\hat{S}}_{ts}|$, where $\mathcal{\hat{S}}_{ts}$ is the subset of test patterns belonging to the non-target class.
We considered 150 target patterns for the training, validation, and test sets, while we used 1500 non-target patterns in the validation and test sets.
As expected, the FPR of the best model is $\simeq 0.1082$, with an overall accuracy of 0.8953.
By means of this interpretation, we could safely affirm that the ``true'' FPR is only $\simeq 0.0082$, since $|\mathcal{S}_{tr}|/|\mathcal{\hat{S}}_{ts}|=0.1$.
The AUC is 0.9884; Fig. \ref{fig:test2_mf} shows the calculated membership values for the test patterns. Although the discrimination is very good, the OCC necessarily commits some mistakes, due the uniform distribution of the non-target patterns.

\begin{figure}[ht!]
 \centering
 \includegraphics[viewport=0 0 461 261,scale=0.5,keepaspectratio=true]{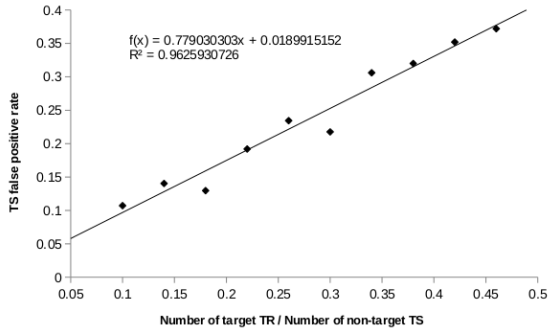}
 \caption{Correlation among $|\mathcal{S}_{tr}|/|\mathcal{\hat{S}}_{ts}|$ and the computed FPR on the test set.}
 \label{fig:regression}
\end{figure}

To demonstrate the reliability of this interpretation of the test, we repeated this experiment by increasing the ratio $|\mathcal{S}_{tr}|/|\mathcal{\hat{S}}_{ts}|$ from 0.1 to 0.475, and, accordingly, increasing also the spread of the target patterns over the domain.
In Fig. \ref{fig:regression} we show the linear correlation between the increments of the $|\mathcal{S}_{tr}|/|\mathcal{\hat{S}}_{ts}|$ ratio and the calculated FPR over the respective test set.
As it was expected, there is a strong linear relationship among those two quantities (correlation coefficient is $\simeq 0.96$), which demonstrates that the implicit FPR obtained with this method of generation of non-target patterns is predictable.

\subsection{Comparison on UCI datasets}
\label{sec:comparison}

\begin{figure}[ht!]
 \centering
 \includegraphics[viewport=0 0 601 451,scale=0.40,keepaspectratio=true]{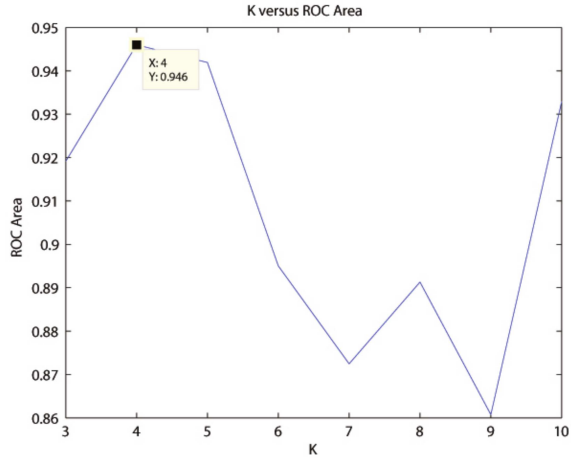}
 \caption{AUC values for different partition orders achieved on the Ecoli (\textbf{E}) dataset.}
 \label{fig:Ecoli_k_search}
\end{figure}

In Tab. \ref{tab:uci_ds} we show the UCI datasets considered in this study; all datasets are standardized with zero mean and unitary variance.
In Tab. \ref{tab:uci_ds_results} we report the results in terms of AUC obtained by our system, denoted as ``OCC{\_}System'', together with those retrieved from the literature (see Ref. \cite{eocc__arxiv} and references therein).
Results in Tab. \ref{tab:uci_ds_results} show how the OCC{\_}System reaches satisfactory performances on the considered five UCI datasets.
Notably, considering \textbf{D} and \textbf{L} datasets the OCC{\_}System achieves the highest AUC.
It is worth pointing out that the proposed OCC achieves a good AUC on the dataset with the highest number of features (i.e., eight for dataset \textbf{D}).
In Fig. \ref{fig:Ecoli_k_search} we show the trend of the AUC achieved for the \textbf{E} dataset with respect to the partition value $k$ ranging in $\{3, ..., 10\}$.
The figure shows how the OCC{\_}System, although it reaches slightly different AUC values varying the $k$ parameter, for some of that, i.e., $k=4$ and $k=10$, the performance is still comparable.
This fact suggests that, at least on the considered dataset, the proposed OCC does not overfit the model reaching good performances also with partitions of lower order.
\begin{table*}[th!]\scriptsize
\begin{center}
\caption{UCI datasets considered in this study.}
\label{tab:uci_ds}
\begin{tabular}{|c|c|c|c|c|c|c|}
\hline
\textbf{UCI Dataset} & \textbf{Acronym} & \textbf{Target class} & \textbf{\# Target} & \textbf{\# Non-target} & \textbf{\# Params} \\
\hline
Biomed & BI & normal & 127 & 67 & 5 \\
Breast Wisconsin & BW & benign & 458 & 241 & 9 \\
Diabetes (prima indians) & D & present & 500 & 268 & 8 \\
Ecoli & E & pp & 52 & 284 & 7 \\
Iris & I & Iris-setosa & 50 & 100 & 4 \\
Liver & L & healthy & 200 & 145 & 6 \\
\hline
\end{tabular}
\end{center}
\end{table*}
\begin{table*}[thp!]\scriptsize
\begin{center}
\caption{Test set results showing the average AUC values together with the standard deviations. Best results are reported in bold. The ``-'' symbol indicates that the result is not available.}
\label{tab:uci_ds_results}
\begin{tabular}{|c|c|c|c|c|c|c|}
\hline
\textbf{System/Dataset}  & \textbf{BI} & \textbf{BW} & \textbf{D} & \textbf{E} & \textbf{I} & \textbf{L} \\
\hline
\rowcolor{lgray} OCC{\_}System & 0.904(0.013) & \textbf{0.996(0.002)} & \textbf{0.756(0.003)} & 0.949(0.008) &  \textbf{1.000(0.000)} &  \textbf{0.652(0.011)} \\
\hline
EOCC-1 &  0.867(0.005) & 0.853(0.020) & 0.670(0.024) & 0.928(0.011) & \textbf{1.000(0.000)} & 0.396(0.016) \\
EOCC-2  & 0.878(0.006) & 0.995(0.001) & 0.751(0.012) & \textbf{0.957(0.004)} & \textbf{1.000(0.000)} & 0.460(0.026) \\
EOCC-2\_10\%  & 0.862(0.016) & 0.995(0.002) & 0.709(0.023) & \textbf{0.954(0.007)} & \textbf{1.000(0.000)} & 0.452(0.026) \\
Gauss  & 0.899(0.005) & 0.985(0.001) & 0.721(0.003) & 0.929(0.003) & \textbf{1.000(0.000)} & 0.509(0.005) \\
MoG  & 0.911(0.008) & 0.984(0.002) & 0.738(0.003) & 0.929(0.003) & \textbf{1.000(0.000)} & 0.494(0.006) \\
Na\"{\i}ve Parzen  & \textbf{0.931(0.002)} & 0.987(0.001) & 0.678(0.003) & 0.930(0.008) & \textbf{1.000(0.000)} & 0.484(0.008) \\
Parzen  & 0.915(0.009) & 0.991(0.001) & 0.756(0.002) & 0.929(0.005) & \textbf{1.000(0.000)} & 0.469(0.008) \\
\textit{k}-Means & 0.902(0.009) & 0.984(0.001) & 0.712(0.010) & 0.878(0.015) & \textbf{1.000(0.000)} & 0.469(0.014) \\
1-NN & 0.914(0.012) & 0.991(0.001) & 0.721(0.002) & 0.906(0.008) & \textbf{1.000(0.000)} & 0.511(0.007) \\
\textit{k}-NN  & 0.914(0.012) & 0.991(0.001) & 0.721(0.002) & 0.906(0.008) & \textbf{1.000(0.000)} & 0.511(0.007) \\
Auto-encoder  & 0.890(0.013) & 0.960(0.002) & 0.658(0.005) &  0.888(0.023) & \textbf{1.000(0.000)} & 0.608(0.008) \\
PCA  & 0.776(0.031) & 0.920(0.004) & 0.640(0.006) & 0.655(0.013) & 0.920(0.008) & 0.608(0.008) \\
SOM  & 0.908(0.006) & 0.990(0.002) & 0.709(0.009) & 0.898(0.004) & \textbf{1.000(0.000)} & 0.487(0.017) \\
MST\_CD  & 0.914(0.012) & 0.992(0.001) & 0.715(0.003) & 0.899(0.009) & \textbf{1.000(0.000)} & - \\
\textit{k}-Centres & 0.906(0.015) & 0.984(0.002) & 0.678(0.009) & 0.870(0.023) & \textbf{1.000(0.000)} & 0.483(0.006) \\
SVDD  & 0.915(0.009) & 0.988(0.001) & 0.732(0.005) & 0.922(0.010) & \textbf{1.000(0.000)} & 0.490(0.010) \\
MPM  & 0.909(0.010) & 0.991(0.001) & 0.729(0.003) & 0.922(0.007) & \textbf{1.000(0.000)} & 0.521(0.011) \\
LPDD  & 0.889(0.008) & 0.989(0.001) & 0.634(0.005) & 0.947(0.004) & \textbf{1.000(0.000)} & 0.506(0.005) \\
\hline
\end{tabular}
\end{center}
\end{table*}

\subsection{Results and discussion on ACEA data}
\label{sec:acea}

\begin{figure}[ht!]
 \centering
 \includegraphics[viewport=0 0 477 375,scale=0.5,keepaspectratio=true]{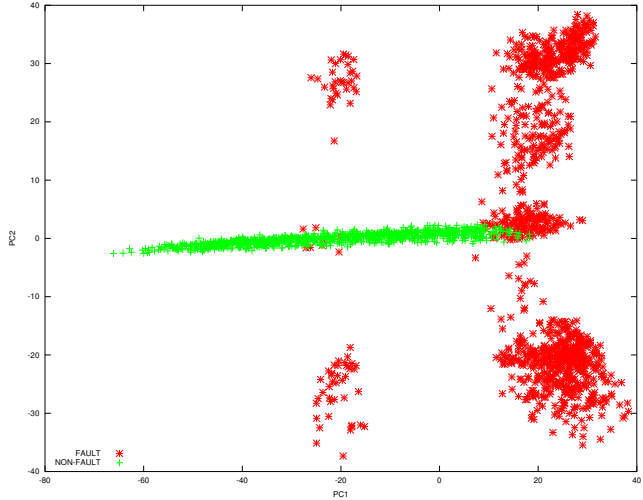}
 \caption{First two components of the PCA elaborated over the dissimilarity matrix constructed from the ACEA dataset containing either fault and non-fault (uniformly generated) patterns.}
 \label{fig:pca_dissmatrix_faults}
\end{figure}

The ACEA dataset available for our experiments does not contain instances of non-fault situations (i.e., normal functioning of the system).
This fact creates some difficulty in evaluating any data-driven inference mechanism.
To generate instances of non-target patterns, we use the method discussed in the second experiment presented in Sec. \ref{sec:synth}.
Non-target patterns are formed by randomly generating, with a uniform distribution, each feature value characterizing a FP (see Sec. \ref{sec:pattern_representation} for details on the features).
Since the dataset contains heterogeneous data types, such a uniform generation mechanism seems to be the most appropriate one. 
In Fig. \ref{fig:pca_dissmatrix_faults} we show the first two components of the PCA calculated over the dissimilarity matrix, $D_{ij}=d(x_i, x_j), \forall x_i,x_j\in\mathcal{S}$, generated for the entire available ACEA dataset $\mathcal{S}$, containing either fault and non-fault instances; $d(\cdot, \cdot)$ in Eq. \ref{eq:Custom_Distance} is computed by using unitary weights.
The herein considered dataset is divided in training, validation, and test sets according to the following splits.
The training set is composed of 532 fault patterns; in the validation set we have 470 fault and 500 non-fault patterns; finally for the test we have 82 fault and 500 non-fault patterns.

Tab. \ref{tab:results_acea} shows the obtained results. We tested the OCC for five different values of \textit{k} in the \textit{k}-means algorithm. For each $k$, we repeated the test five different times by changing the random seed of the pseudo-random number generator driving the stochastic behaviors of the \textit{k}-means initialization procedure and of the genetic algorithm. Hence the results are intended as averages with related standard deviations.
In the table we show the false positive rate (FPR), the recall (R), the accuracy (A), the area under the ROC curve (AUC), the fuzzy entropy (FE) of the winning model, and finally the mutual information (MI) among the fitness of the best individual and the estimated entropy of the related weights. Let us focus now on the first five columns.
Results in terms of recognition, considering both hard and soft decisions, are in general very good.
The best overall result is obtained with $k=7$ (please note a somewhat clear 7-cluster structure of the fault patterns appearing from the first two components of the PCA in Fig. \ref{fig:pca_dissmatrix_faults}). Notably, FPRs are always very low, demonstrating the capability of the proposed OCC of synthesizing well-defined and effective decision regions.
Notwithstanding non-fault patterns are generated by using a uniform distribution over each feature describing a FP (please see Fig. \ref{fig:pca_dissmatrix_faults}), the system is able to isolate such patterns correctly.
On the same line of thoughts, both accuracy and AUC are nearly one, denoting an almost perfect recognition of faults.
The FE of the best models (see Sec. \ref{sec:class_reliability}) is almost negligible (in accord with the very high AUC), which tells us that the soft decisions are also highly reliable.

\begin{figure}[ht!]
\centering

\subfigure[AUC and Accuracy (A).]{
\includegraphics[viewport=0 0 352 243,scale=0.6,keepaspectratio=true]{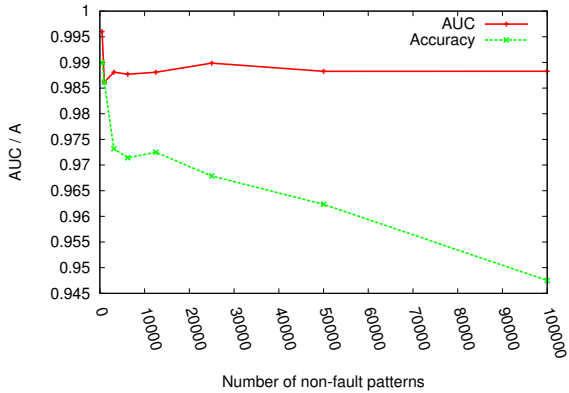}
\label{fig:auc_a}}
~
\subfigure[FPR.]{
\includegraphics[viewport=0 0 352 243,scale=0.6,keepaspectratio=true]{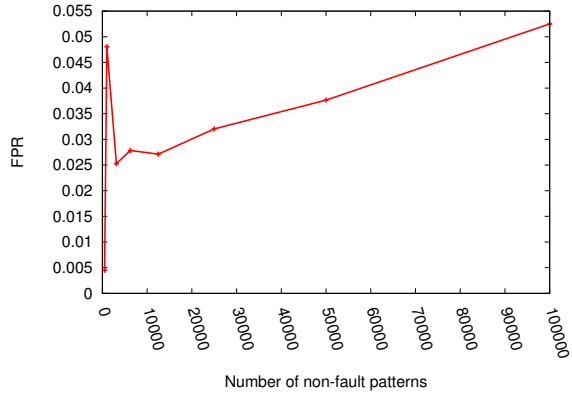}
\label{fig:fpr}}

\caption{Test set results by increasing the number of considered non-fault patterns in the test set. Fig. \ref{fig:auc_a} shows the AUC and A while Fig. \ref{fig:fpr} shows the FPR.}
\label{fig:auc_a_fpr}
\end{figure}

To offer an argument to demonstrate the validity of the herein shown results, and accordingly with the generation mechanism for the non-fault patterns, in Fig. \ref{fig:auc_a_fpr} we show the performances obtained by progressively increasing the number of non-fault patterns. The generation mechanism of course remains the same. In the figure we show the best performing configuration of the system for each non-fault patterns set size.
From Fig. \ref{fig:auc_a} it is possible to deduce that the AUC does not vary significantly by increasing the number of non-fault patterns.
On the other hand, the accuracy (A) is slightly more affected, although still denoting good results. This could be interpreted as a sign of robustness of the fuzzy set based soft decision mechanism herein proposed.
Finally, Fig. \ref{fig:fpr} reports the FPR that denotes a trend correlated with the one of the AUC.
\begin{table}[thp!]\scriptsize
\begin{center}
\caption{Average test sets results on the ACEA data. Results are reported for five different values of \textit{k} for the \textit{k}-means algorithm.}
\label{tab:results_acea}
\begin{tabular}{|c|c|c|c|c|c|c|}
\hline
\textbf{K} & \textbf{FPR} & \textbf{R} & \textbf{A} & \textbf{AUC} & \textbf{FE} & \textbf{MI} \\
\hline
4 & 0.00449\textpm0.003 & 0.94200\textpm0.027 & 0.98800\textpm0.005 & 0.99600\textpm0.002 & 0.00511\textpm0.002 & 0.70529\textpm0.059 \\
\hline
5 & 0.00599\textpm0.001 & 0.95700\textpm0.012 & 0.98900\textpm0.001 & 0.99500\textpm0.001 & 0.00483\textpm0.002 & 0.61681\textpm0.082 \\
 \hline
6 & 0.00649\textpm0.002 & 0.93600\textpm0.020 & 0.98500\textpm0.002 & 0.99500\textpm0.001 & 0.00497\textpm0.001 & 0.65527\textpm0.029 \\
 \hline
7 & 0.00449\textpm0.003 & 0.95700\textpm0.015 & 0.99000\textpm0.003 & 0.99600\textpm0.002 & 0.00155\textpm0.018 & 0.78783\textpm0.072 \\
\hline
8 & 0.00499\textpm0.003 & 0.95700\textpm0.007 & 0.99000\textpm0.003 & 0.99500\textpm0.003 & 0.00165\textpm0.014 & 0.72686\textpm0.131 \\
\hline
\end{tabular}
\end{center}
\end{table}

\begin{figure}[ht!]
 \centering
 \includegraphics[viewport=0 0 842 500,scale=0.35,keepaspectratio=true]{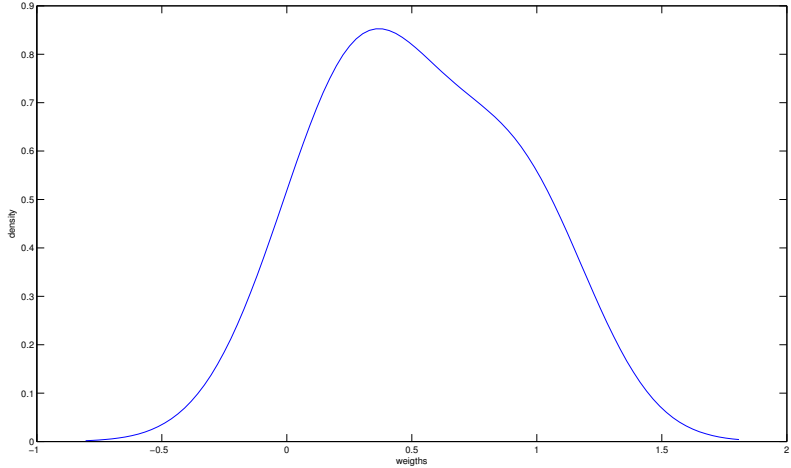}
 \caption{Density reconstructed for the best solution -- the weights of the dissimilarity measure (\ref{eq:Custom_Distance}) -- found by the OCC with $k=7$. The density is not uniform over the $[0, 1]$ range.}
 \label{fig:density_weigth_k7}
\end{figure}

Let us now go back to discuss the information conveyed by the MI column in Tab. \ref{tab:results_acea}.
As it is clear from Eq. \ref{eq:Custom_Distance}, each feature is weighted by a specific $w_j\in[0, 1]$. Such weights, $\underline{\mathbf{w}}$, are calculated by exploiting the fitness function in Eq. \ref{eq:obj_kmeans}, which searches for the best-performing parameters configuration on a suitable validation set.
It is important to assess the importance of the considered features describing a fault pattern with respect to the classification problem at hand. Notably, we expect to find features that are, in average, more relevant than others, thus contributing with more impact in the discrimination process.
To verify such a hypothesis, in Fig. \ref{fig:density_weigth_k7} we show the estimated density (kernel-based estimator) of the weights related to the best-performing individual; $k=7$ is considered here.
From the figure it is possible to deduce that the features are not uniformly weighted.

It is worth noting that, regardless of the value of $k$, there is a subset of features that is always associated with high weights (details not shown). Such features are: ``Time start'', ``Primary station fault distance'', ``Median point'', ``Max temperature'', ``Rain'', ``Interruption (breaker)'', and ``Petersen alarms''.
Such features confirm what the expertise of the ACEA company indicates as most important factors congruent to a LF.
The time of the day (``Time start'') is an important variable due to the changing on energy demand that normally presents two peaks, one at the middle of the day and one in the later evening.
The distance from the PS (``Primary station fault distance'') and the absolute position of the LF (``Median point'') is also a characterizing property, confirming the hypothesis on the amount of electric current that flow along a backbone -- see Sec. \ref{sec:Spatial_Data}. 
Other important indicators advised by the ACEA company experts are the weather conditions, specially the millimeters of rains and the maximum temperature in a day. In fact, it is well-known (and it is also reasonable) that the events of heavy rain are strongly correlated with grid black-outs.
Accordingly, a strong discrimination is conveyed by the sequences of automatically registered events. In particular, the interruptions registered in the PSs due to the opening of the prevention breakers caused by short circuits on the power line and the Petersen alarms, which are registered as soon as a loss of the dielectric capacity of the power equipments is detected.
This phenomena can be physically characterized by overheating (due to the high currents) in which the dielectric of power equipments changes his properties for a short period.
The presence of bursts of these events could be indicative of an imminent LF. This last perspective will be studied in detail in future research studies.

\begin{figure}[ht!]
 \centering
 \includegraphics[viewport=0 0 410 302,scale=0.7,keepaspectratio=true]{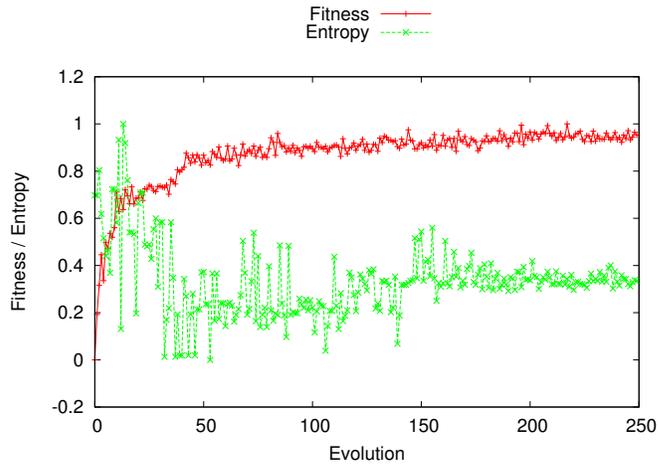}
 \caption{Sequences of fitness values and entropy estimations on the weights corresponding to the best individual solution found at each evolution of the OCC model optimization.}
 \label{fig:mi}
\end{figure}

The OCC system is trained by cross-validation, exploiting the fitness function in Eq. \ref{eq:obj_kmeans}.
Here we aim to demonstrate that the solutions found by the OCC become more informative as the system finds better solutions, i.e., with solutions that improve the discrimination of faults/non-faults.
This result helps us in justifying the claim that the final solutions found by the proposed OCC are informative (the considered features have different importance in the discrimination process).
As previously described, the synthesis of the OCC consists in performing a certain number of iterations until the stop criterion in reached.
An important observation is that, during the iterations characterizing the optimization, the fitness of the best-performing individual has a non-decreasing trend (this is obtained since we use a form of elitism in our GA). Accordingly, we expect to observe a non-decreasing trend also for what concerns the estimated entropy of the distribution underlying the weights $\underline{\mathbf{w}}$ related to the best-performing individual of each iteration.
To demonstrate such a claim, we calculated the (non-linear) correlation among the sequence of fitness values and the sequence of estimated entropy values, both related to the best-performing solutions found at each iteration/evolution.
Fig. \ref{fig:mi} shows those two sequences for a specific test performed with $k=7$. Although at the beginning the two series are not very correlated, they stabilize to a similar trend that is captured via the estimation of the mutual information \cite{Moon_MI__2002}.
The MI column in Tab. \ref{tab:results_acea} reports hence the average mutual information estimated between those two series for each $k$; MI values fall within the $[0, 1]$ range.
As it is possible to observe, the non-linear correlation is in general good, especially in the $k=7$ case, where also the MI reaches its maximum score.

\section{Conclusions}
\label{sec:conclusions}

Predicting faults in real-world smart grid systems is a challenging task. This is due to the high variability of those types of systems and also to the heterogeneity of the data that actually characterize a fault situation.
In our study, we modeled localized faults in the ACEA smart grid by means of several heterogeneous features.
The proposed one-class classifier is based on an interplay among clustering and dissimilarity measure learning techniques, where specialized measure have been designed to deal with each feature type (e.g., categorical, metric, and time series). The classifier synthesis is guided by a genetic algorithm, which is in charge to optimize the weighting parameters of the dissimilarity measure adopted in the input domain, as well as the tolerances defining decision region boundaries.
The proposed system is able to provide both hard (i.e., Boolean) and soft decisions regarding the recognition of a test pattern.
Soft decisions are introduced also to offer an overall measure of reliability concerning the decisions; non reliable decisions can be individuated by calculating the fuzzy entropy of the resulting membership function.
Experimental evaluations performed on the ACEA data demonstrate the effectiveness of the proposed solution in a real-world smart grid setting.

Future directions include the possibility to evaluate other clustering algorithms and different global optimization schemes.
Particular interest will be given to the study of the time series of electrical current values measured in different areas of the ACEA power grid, aiming to characterize the underlying (dynamic) system generating such data with the aim to find correlation/causation rules with the observed faults.

\bibliographystyle{abbrvnat}
\bibliography{Bibliography}

\begin{thebibliography}{57}
\providecommand{\natexlab}[1]{#1}
\providecommand{\url}[1]{\texttt{#1}}
\expandafter\ifx\csname urlstyle\endcsname\relax
  \providecommand{\doi}[1]{doi: #1}\else
  \providecommand{\doi}{doi: \begingroup \urlstyle{rm}\Url}\fi

\bibitem[ACE()]{ACEA_SG_Pilot_Proj}
the acea smart grid pilot project (in italian).
\newblock URL
  \url{http://www.autorita.energia.it/allegati/operatori/elettricita/smartgrid/V%20Rel%20smart%20ACEA%20D.pdf}.

\bibitem[Doe()]{Doe:2013:Online}
{CEI} - comitato elettrotecnico italiano.
\newblock URL \url{http://www.ceiweb.it/it/}.

\bibitem[Ene()]{Energy_Information_Administration}
International energy outlook 2011 - energy information administration.
\newblock URL \url{http://www.eia.gov/forecasts/ieo/index.cfm}.

\bibitem[sma()]{smartgrids_ETP}
The {SmartGrids} european technology platform {\textbar} {SmartGrids}.
\newblock URL \url{http://www.smartgrids.eu/ETPSmartGrids}.

\bibitem[Ace(2010)]{Acea_Road_mapr_2010}
{La Road Map per la continuità del servizio di distrbuzione elettrica di Roma,
  Acea Distribuzione SpA, Aracne editrice, Città Ducale (RI)}, 2010.

\bibitem[Abdelaziz et~al.(2009)Abdelaziz, Mohammed, Mekhamer, and
  Badr]{abdelaziz2009distribution}
A.~Y. Abdelaziz, F.~M. Mohammed, S.~F. Mekhamer, and M.~A.~L. Badr.
\newblock Distribution systems reconfiguration using a modified particle swarm
  optimization algorithm.
\newblock \emph{Electric Power Systems Research}, 79\penalty0 (11):\penalty0
  1521--1530, 2009.

\bibitem[Abido et~al.(2014)Abido, El-Alfy, and Sheraz]{abido2014computational}
M.~A. Abido, E.-S.~M. El-Alfy, and M.~Sheraz.
\newblock Computational intelligence in smart grids: Case studies.
\newblock In \emph{Computational Intelligence for Decision Support in
  Cyber-Physical Systems}, pages 265--292. Springer, 2014.

\bibitem[Afzal and Pothamsetty(2012)]{6175733}
M.~Afzal and V.~Pothamsetty.
\newblock Analytics for distributed smart grid sensing.
\newblock In \emph{Innovative Smart Grid Technologies (ISGT), 2012 IEEE PES},
  pages 1--7, 2012.
\newblock \doi{10.1109/ISGT.2012.6175733}.

\bibitem[Amin and Wollenberg(2005)]{amin2005toward}
S.~M. Amin and B.~F. Wollenberg.
\newblock Toward a smart grid: power delivery for the 21st century.
\newblock \emph{IEEE Power and Energy Magazine}, 3\penalty0 (5):\penalty0
  34--41, 2005.

\bibitem[Anderson et~al.(2011)Anderson, Boulanger, Powell, and Scott]{5768096}
R.~Anderson, A.~Boulanger, W.~Powell, and W.~Scott.
\newblock Adaptive stochastic control for the smart grid.
\newblock \emph{Proceedings of the IEEE}, 99\penalty0 (6):\penalty0 1098--1115,
  2011.
\newblock ISSN 0018-9219.
\newblock \doi{10.1109/JPROC.2011.2109671}.

\bibitem[{Anvari Moghaddam} and Seifi(2011)]{6111641}
A.~{Anvari Moghaddam} and A.~R. Seifi.
\newblock Study of forecasting renewable energies in smart grids using linear
  predictive filters and neural networks.
\newblock \emph{Renewable Power Generation, IET}, 5\penalty0 (6):\penalty0
  470--480, 2011.
\newblock ISSN 1752-1416.
\newblock \doi{10.1049/iet-rpg.2010.0104}.

\bibitem[Bache and Lichman(2013)]{Bache+Lichman:2013}
K.~Bache and M.~Lichman.
\newblock {{UCI} Machine Learning Repository}, 2013.
\newblock URL \url{http://archive.ics.uci.edu/ml}.

\bibitem[Bellet et~al.(2013)Bellet, Habrard, and Sebban]{Metric_learning}
A.~Bellet, A.~Habrard, and M.~Sebban.
\newblock A survey on metric learning for feature vectors and structured data.
\newblock \emph{CoRR}, abs/1306.6709, 2013.
\newblock URL \url{http://arxiv.org/abs/1306.6709}.

\bibitem[Cai and Chow(2009)]{5275689}
Y.~Cai and M.-Y. Chow.
\newblock Exploratory analysis of massive data for distribution fault diagnosis
  in smart grids.
\newblock In \emph{Power Energy Society General Meeting, 2009. PES '09. IEEE},
  pages 1--6, 2009.
\newblock \doi{10.1109/PES.2009.5275689}.

\bibitem[{De Santis} et~al.(2013){De Santis}, Rizzi, Sadeghian, and {Frattale
  Mascioli}]{de2013genetic}
E.~{De Santis}, A.~Rizzi, A.~Sadeghian, and F.~M. {Frattale Mascioli}.
\newblock Genetic optimization of a fuzzy control system for energy flow
  management in micro-grids.
\newblock In \emph{2013 Joint IFSA World Congress and NAFIPS Annual Meeting},
  pages 418--423. IEEE, 2013.

\bibitem[{De Santis} et~al.(2014){De Santis}, Livi, Sadeghian, and
  Rizzi]{enrico_occ}
E.~{De Santis}, L.~Livi, A.~Sadeghian, and A.~Rizzi.
\newblock Fault recognition in smart grids by a one-class classification
  approach.
\newblock In \emph{{Proceedings of the 2014 International Joint Conference on
  Neural Networks}}, pages 1949--1956. IEEE, July 2014.
\newblock \doi{10.1109/IJCNN.2014.6889668}.

\bibitem[{Del Vescovo} et~al.(2014){Del Vescovo}, Livi, {Frattale Mascioli},
  and Rizzi]{delvescovo+livi+rizzi+frattalemascioli2011}
G.~{Del Vescovo}, L.~Livi, F.~M. {Frattale Mascioli}, and A.~Rizzi.
\newblock {On the Problem of Modeling Structured Data with the MinSOD
  Representative}.
\newblock \emph{International Journal of Computer Theory and Engineering},
  6\penalty0 (1):\penalty0 9--14, 2014.
\newblock ISSN 1793-8201.
\newblock \doi{10.7763/IJCTE.2014.V6.827}.

\bibitem[Ding et~al.(2014)Ding, Li, Belatreche, and Maguire]{Ding2014313}
X.~Ding, Y.~Li, A.~Belatreche, and L.~P. Maguire.
\newblock An experimental evaluation of novelty detection methods.
\newblock \emph{Neurocomputing}, 135\penalty0 (0):\penalty0 313 -- 327, 2014.
\newblock ISSN 0925-2312.
\newblock \doi{http://dx.doi.org/10.1016/j.neucom.2013.12.002}.

\bibitem[D{\"o}rfler et~al.(2013)D{\"o}rfler, Chertkov, and
  Bullo]{dorfler2013synchronization}
F.~D{\"o}rfler, M.~Chertkov, and F.~Bullo.
\newblock Synchronization in complex oscillator networks and smart grids.
\newblock \emph{Proceedings of the National Academy of Sciences}, 110\penalty0
  (6):\penalty0 2005--2010, 2013.

\bibitem[Fawcett(2006)]{Fawcett:2006:IRA:1159473.1159475}
T.~Fawcett.
\newblock {An Introduction to ROC Analysis}.
\newblock \emph{Pattern Recognition Letters}, 27\penalty0 (8):\penalty0
  861--874, June 2006.
\newblock ISSN 0167-8655.
\newblock \doi{10.1016/j.patrec.2005.10.010}.

\bibitem[Gambardella et~al.(2010)Gambardella, Giacinto, Migliaccio, and
  Montali]{oilspill__2010}
A.~Gambardella, G.~Giacinto, M.~Migliaccio, and A.~Montali.
\newblock One-class classification for oil spill detection.
\newblock \emph{Pattern Analysis and Applications}, 13\penalty0 (3):\penalty0
  349--366, 2010.
\newblock ISSN 1433-7541.
\newblock \doi{10.1007/s10044-009-0164-z}.

\bibitem[Guikema et~al.(2006)Guikema, Davidson, and Haibin]{1645199}
S.~D. Guikema, R.~A. Davidson, and L.~Haibin.
\newblock Statistical models of the effects of tree trimming on power system
  outages.
\newblock \emph{IEEE Transactions on Power Delivery}, 21\penalty0 (3):\penalty0
  1549--1557, 2006.
\newblock ISSN 0885-8977.
\newblock \doi{10.1109/TPWRD.2005.860238}.

\bibitem[Jain(2010)]{Jain:2010:DCY:1755267.1755654}
A.~K. Jain.
\newblock {Data clustering: 50 years beyond K-means}.
\newblock \emph{Pattern Recognition Letters}, 31\penalty0 (8):\penalty0
  651--666, June 2010.
\newblock ISSN 0167-8655.
\newblock \doi{10.1016/j.patrec.2009.09.011}.

\bibitem[Juszczak et~al.(2009)Juszczak, Tax, P\c{e}kalska, and
  Duin]{Juszczak20091859}
P.~Juszczak, D.~M.~J. Tax, E.~P\c{e}kalska, and R.~P.~W. Duin.
\newblock Minimum spanning tree based one-class classifier.
\newblock \emph{Neurocomputing}, 72\penalty0 (7--9):\penalty0 1859 -- 1869,
  2009.
\newblock ISSN 0925-2312.
\newblock \doi{http://dx.doi.org/10.1016/j.neucom.2008.05.003}.

\bibitem[Kemmler et~al.(2013{\natexlab{a}})Kemmler, Rodner, Rösch, Popp, and
  Denzler]{Kemmler201329}
M.~Kemmler, E.~Rodner, P.~Rösch, J.~Popp, and J.~Denzler.
\newblock Automatic identification of novel bacteria using {R}aman spectroscopy
  and gaussian processes.
\newblock \emph{Analytica Chimica Acta}, 794:\penalty0 29 -- 37,
  2013{\natexlab{a}}.
\newblock ISSN 0003-2670.
\newblock \doi{http://dx.doi.org/10.1016/j.aca.2013.07.051}.

\bibitem[Kemmler et~al.(2013{\natexlab{b}})Kemmler, Rodner, Wacker, and
  Denzler]{Kemmler2013}
M.~Kemmler, E.~Rodner, E.-S. Wacker, and J.~Denzler.
\newblock One-class classification with gaussian processes.
\newblock \emph{Pattern Recognition}, 46\penalty0 (12):\penalty0 3507 -- 3518,
  2013{\natexlab{b}}.
\newblock ISSN 0031-3203.
\newblock \doi{http://dx.doi.org/10.1016/j.patcog.2013.06.005}.

\bibitem[Khan and Madden(2010)]{one-class_survey__2010}
S.~S. Khan and M.~G. Madden.
\newblock A survey of recent trends in one class classification.
\newblock In L.~Coyle and J.~Freyne, editors, \emph{Artificial Intelligence and
  Cognitive Science}, volume 6206 of \emph{Lecture Notes in Computer Science},
  pages 188--197. Springer Berlin Heidelberg, 2010.
\newblock ISBN 978-3-642-17079-9.
\newblock \doi{10.1007/978-3-642-17080-5_21}.

\bibitem[Li et~al.(2010)Li, Qiao, Sun, Wan, Wang, Xia, Xu, and Zhang]{5535240}
F.~Li, W.~Qiao, H.~Sun, H.~Wan, J.~Wang, Y.~Xia, Z.~Xu, and P.~Zhang.
\newblock Smart transmission grid: Vision and framework.
\newblock \emph{IEEE Transactions on Smart Grid}, 1\penalty0 (2):\penalty0
  168--177, Sept 2010.
\newblock ISSN 1949-3053.
\newblock \doi{10.1109/TSG.2010.2053726}.

\bibitem[Livi and Rizzi(2013)]{Livi_ga_2013}
L.~Livi and A.~Rizzi.
\newblock {Graph ambiguity}.
\newblock \emph{Fuzzy Sets and Systems}, 221\penalty0 (0):\penalty0 24--47,
  2013.
\newblock ISSN 0165-0114.
\newblock \doi{10.1016/j.fss.2013.01.001}.

\bibitem[Livi et~al.(2014)Livi, Sadeghian, and Pedrycz]{eocc__arxiv}
L.~Livi, A.~Sadeghian, and W.~Pedrycz.
\newblock Entropic one-class classifiers.
\newblock \emph{ArXiv preprint arxiv:1407.7556}, Jul 2014.

\bibitem[Machowski et~al.(2011)Machowski, Bialek, and
  Bumby]{machowski2011power}
J.~Machowski, J.~Bialek, and J.~Bumby.
\newblock \emph{Power system dynamics: stability and control}.
\newblock John Wiley \& Sons, 2011.

\bibitem[Mei et~al.(2011)Mei, Cao, and Zhang]{mei_power_2011}
S.~Mei, M.~Cao, and X.~Zhang.
\newblock \emph{Power grid complexity}.
\newblock Springer, 2011.

\bibitem[Molderink et~al.(2010)Molderink, Bakker, Bosman, Hurink, and
  Smit]{molderink2010management}
A.~Molderink, V.~Bakker, M.~G.~C. Bosman, J.~L. Hurink, and G.~J.~M. Smit.
\newblock Management and control of domestic smart grid technology.
\newblock \emph{IEEE Transactions on Smart Grid}, 1\penalty0 (2):\penalty0
  109--119, 2010.

\bibitem[Moon et~al.(1995)Moon, Rajagopalan, and Lall]{Moon_MI__2002}
Y.-I. Moon, B.~Rajagopalan, and U.~Lall.
\newblock {Estimation of mutual information using kernel density estimators}.
\newblock \emph{Physical Review E}, 52:\penalty0 2318--2321, Sep 1995.
\newblock \doi{10.1103/PhysRevE.52.2318}.

\bibitem[M\"{u}ller(2007)]{muller_dtw}
M.~M\"{u}ller.
\newblock Dynamic time warping.
\newblock In \emph{Information Retrieval for Music and Motion}, pages 69--84.
  Springer Berlin Heidelberg, 2007.
\newblock ISBN 978-3-540-74047-6.
\newblock \doi{10.1007/978-3-540-74048-3_4}.

\bibitem[Park and Jun(2009)]{Park20093336}
H.-S. Park and C.-H. Jun.
\newblock A simple and fast algorithm for k-medoids clustering.
\newblock \emph{Expert Systems with Applications}, 36\penalty0 (2, Part
  2):\penalty0 3336 -- 3341, 2009.
\newblock ISSN 0957-4174.
\newblock \doi{http://dx.doi.org/10.1016/j.eswa.2008.01.039}.
\newblock URL
  \url{http://www.sciencedirect.com/science/article/pii/S095741740800081X}.

\bibitem[P\c{e}kalska et~al.(2002)P\c{e}kalska, Tax, and Duin]{NIPS2002_2163}
E.~P\c{e}kalska, D.~Tax, and R.~P.~W. Duin.
\newblock One-class lp classifiers for dissimilarity representations.
\newblock In S.~Becker, S.~Thrun, and K.~Obermayer, editors, \emph{Advances in
  Neural Information Processing Systems}, volume~15, pages 761--768. 2002.

\bibitem[Pedrycz and Gomide(1998)]{pedrycz1998introduction}
W.~Pedrycz and F.~Gomide.
\newblock \emph{{An Introduction to Fuzzy Sets: Analysis and Design}}.
\newblock {Complex Adaptive Systems}. Mit Press, 1998.
\newblock ISBN 9780262161718.

\bibitem[Raheja et~al.(2006)Raheja, Llinas, Nagi, and
  Romanowski]{Raheja2006-J-IJPR}
D.~Raheja, J.~Llinas, R.~Nagi, and C.~Romanowski.
\newblock {Data fusion/data mining-based architecture for condition-based
  maintenance}.
\newblock \emph{International Journal of Production Research}, 44\penalty0
  (14):\penalty0 2869--2887, July 2006.
\newblock ISSN 0020-7543.
\newblock \doi{10.1080/00207540600654509}.

\bibitem[Rizzi et~al.(2013)Rizzi, Livi, Tahayori, and
  Sadeghian]{t2vsdiss__ifsanafips2013}
A.~Rizzi, L.~Livi, H.~Tahayori, and A.~Sadeghian.
\newblock {Matching general type-2 fuzzy sets by comparing the vertical
  slices}.
\newblock In \emph{{2013 Joint IFSA World Congress and NAFIPS Annual Meeting
  (IFSA/NAFIPS)}}, pages 866--871, 2013.
\newblock \doi{10.1109/IFSA-NAFIPS.2013.6608514}.

\bibitem[Saha et~al.(2011)Saha, Aldeen, and Tan]{Saha2011887}
S.~Saha, M.~Aldeen, and C.~P. Tan.
\newblock Fault detection in transmission networks of power systems.
\newblock \emph{International Journal of Electrical Power \& Energy Systems},
  33\penalty0 (4):\penalty0 887 -- 900, 2011.
\newblock ISSN 0142-0615.
\newblock \doi{http://dx.doi.org/10.1016/j.ijepes.2010.12.026}.

\bibitem[Sch\"olkopf et~al.(2000)Sch\"olkopf, Williamson, Smola, Shawe-Taylor,
  and Platt]{SchWilSmoShaetal00}
B.~Sch\"olkopf, R.~Williamson, A.~Smola, J.~Shawe-Taylor, and J.~Platt.
\newblock Support vector method for novelty detection.
\newblock In \emph{Neural Information Processing Systems}, pages 582--588,
  2000.

\bibitem[Shahid et~al.(2012)Shahid, Aleem, Naqvi, and Zaffar]{6477812}
N.~Shahid, S.~Aleem, I.~Naqvi, and N.~Zaffar.
\newblock Support vector machine based fault detection amp; classification in
  smart grids.
\newblock In \emph{Globecom Workshops (GC Wkshps), 2012 IEEE}, pages
  1526--1531, 2012.
\newblock \doi{10.1109/GLOCOMW.2012.6477812}.

\bibitem[Shanker and Rajagopalan(2007)]{PiyushShanker20071407}
A.~P. Shanker and A.~Rajagopalan.
\newblock Off-line signature verification using {DTW}.
\newblock \emph{Pattern Recognition Letters}, 28\penalty0 (12):\penalty0 1407
  -- 1414, 2007.
\newblock ISSN 0167-8655.
\newblock \doi{10.1016/j.patrec.2007.02.016}.

\bibitem[Storti et~al.(2013)Storti, Possemato, Paschero, Rizzi, and
  Mascioli]{6608435}
G.~Storti, F.~Possemato, M.~Paschero, A.~Rizzi, and F.~Mascioli.
\newblock Optimal distribution feeders configuration for active power losses
  minimization by genetic algorithms.
\newblock In \emph{IFSA World Congress and NAFIPS Annual Meeting (IFSA/NAFIPS),
  2013 Joint}, pages 407--412, 2013.
\newblock \doi{10.1109/IFSA-NAFIPS.2013.6608435}.

\bibitem[Sweetser(2011)]{6039785}
C.~L. Sweetser.
\newblock The importance of advanced diagnostic methods for higher availability
  of power transformers and ancillary components in the era of smart grid.
\newblock In \emph{Power and Energy Society General Meeting, 2011 IEEE}, pages
  1--3, 2011.
\newblock \doi{10.1109/PES.2011.6039785}.

\bibitem[Tax and Duin(1999)]{Tax19991191}
D.~M.~J. Tax and R.~P.~W. Duin.
\newblock Support vector domain description.
\newblock \emph{Pattern Recognition Letters}, 20\penalty0 (11–13):\penalty0
  1191 -- 1199, 1999.
\newblock ISSN 0167-8655.
\newblock \doi{http://dx.doi.org/10.1016/S0167-8655(99)00087-2}.

\bibitem[Utkin(2012)]{Utkin:2012:FOC:2213741.2433967}
L.~V. Utkin.
\newblock Fuzzy one-class classification model using contamination
  neighborhoods.
\newblock \emph{Advances in Fuzzy Systems}, 2012:\penalty0 22, 2012.
\newblock \doi{10.1155/2012/984325}.

\bibitem[Venayagamoorthy(2009)]{Venayagamoorthy__2009}
G.~K. Venayagamoorthy.
\newblock {Potentials and promises of computational intelligence for smart
  grids}.
\newblock In \emph{Power Energy Society General Meeting, 2009. PES '09. IEEE},
  pages 1--6, July 2009.
\newblock \doi{10.1109/PES.2009.5275224}.

\bibitem[Venayagamoorthy(2011)]{5952102}
G.~K. Venayagamoorthy.
\newblock Dynamic, stochastic, computational, and scalable technologies for
  smart grids.
\newblock \emph{IEEE Computational Intelligence Magazine}, 6\penalty0
  (3):\penalty0 22--35, 2011.
\newblock ISSN 1556-603X.
\newblock \doi{10.1109/MCI.2011.941588}.

\bibitem[Vincenty(1975)]{vincenty_direct_1975}
T.~Vincenty.
\newblock {Direct and inverse solutions of geodesics on the ellipsoid with
  application of nested equations}.
\newblock \emph{Survey Review}, XXII, Apr. 1975.

\bibitem[Wang and Lai(2013)]{Wang2013875}
C.-D. Wang and J.~Lai.
\newblock Position regularized support vector domain description.
\newblock \emph{Pattern Recognition}, 46\penalty0 (3):\penalty0 875 -- 884,
  2013.
\newblock ISSN 0031-3203.
\newblock \doi{http://dx.doi.org/10.1016/j.patcog.2012.09.018}.

\bibitem[Wang and Zhao(2009)]{5234528}
Z.~Wang and P.~Zhao.
\newblock Fault location recognition in transmission lines based on support
  vector machines.
\newblock In \emph{Computer Science and Information Technology, 2009. ICCSIT
  2009. 2nd IEEE International Conference on}, pages 401--404, 2009.
\newblock \doi{10.1109/ICCSIT.2009.5234528}.

\bibitem[Werbos(2009)]{5179088}
P.~J. Werbos.
\newblock Putting more brain-like intelligence into the electric power grid:
  What we need and how to do it.
\newblock In \emph{Proceedings of the International Joint Conference on Neural
  Networks}, pages 3356--3359, 2009.
\newblock \doi{10.1109/IJCNN.2009.5179088}.

\bibitem[Wetzer(2005)]{1600559}
J.~Wetzer.
\newblock Maintaining future (electrical) power systems.
\newblock In \emph{Future Power Systems, 2005 International Conference on},
  pages 6 pp.--6, 2005.
\newblock \doi{10.1109/FPS.2005.204286}.

\bibitem[Zhang et~al.(2009)Zhang, Liu, Wang, and Wang]{5156572}
Y.~Zhang, Y.~Liu, X.~Wang, and Z.~Wang.
\newblock Fault pattern recognition in power system engineering.
\newblock In \emph{Industrial Mechatronics and Automation, 2009. ICIMA 2009.
  International Conference on}, pages 109--112, 2009.
\newblock \doi{10.1109/ICIMA.2009.5156572}.

\bibitem[Zhang et~al.(2011)Zhang, Wang, Zhang, and Ma]{Zhang2011791}
Y.-G. Zhang, Z.-P. Wang, J.-F. Zhang, and J.~Ma.
\newblock Fault localization in electrical power systems: A pattern recognition
  approach.
\newblock \emph{International Journal of Electrical Power \& Energy Systems},
  33\penalty0 (3):\penalty0 791 -- 798, 2011.
\newblock ISSN 0142-0615.
\newblock \doi{http://dx.doi.org/10.1016/j.ijepes.2011.01.018}.

\end{thebibliography}
\end{document}